\documentclass{article}

\usepackage{microtype}
\usepackage{graphicx}
\usepackage{subcaption}
\usepackage{booktabs} 

\usepackage{hyperref}

\usepackage[preprint]{icml2026}

\usepackage{amsmath}
\usepackage{amssymb}
\usepackage{mathtools}
\usepackage{amsthm}

\usepackage{multirow}
\usepackage[normalem]{ulem}
\usepackage{makecell}
\usepackage{longtable}
\usepackage{listings}
\usepackage[table,xcdraw]{xcolor}

\useunder{\uline}{\ul}{}

\newcommand{\method}{DeltaMem}

\usepackage[most]{tcolorbox}
\newtcolorbox[list inside=prompt,auto counter]{prompt}[1][]{
    enhanced,
    breakable,
    colbacktitle=black!60,
    coltitle=white,
    fontupper=\footnotesize,
    boxsep=5pt,
    left=0pt,
    right=0pt,
    top=0pt,
    bottom=0pt,
    boxrule=1pt,
    #1,
}

\usepackage[capitalize,noabbrev]{cleveref}

\theoremstyle{plain}
\newtheorem{theorem}{Theorem}[section]

\theoremstyle{definition}
\newtheorem{definition}[theorem]{Definition}

\theoremstyle{remark}

\usepackage[textsize=tiny]{todonotes}

\icmltitlerunning{DeltaMem: Towards Agentic Memory Management via Reinforcement Learning}

\begin{document}

\twocolumn[
  \icmltitle{DeltaMem: Towards Agentic Memory Management via Reinforcement Learning}



  \icmlsetsymbol{equal}{*}

  \begin{icmlauthorlist}
    \icmlauthor{Qi Zhang}{ali,zju}
    \icmlauthor{Shen Huang}{ali}
    \icmlauthor{Chu Liu}{ali}
    \icmlauthor{Shouqing Yang}{zju}
    \icmlauthor{Junbo Zhao}{zju}
    \icmlauthor{Haobo Wang}{zju}
    \icmlauthor{Pengjun Xie}{ali}
  \end{icmlauthorlist}

  \icmlaffiliation{ali}{Alibaba Tongyi Lab, Hangzhou, China}
  \icmlaffiliation{zju}{Zhejiang University, Hangzhou, China}

  \icmlcorrespondingauthor{Qi Zhang}{cheung\_se@zju.edu.cn}
  \icmlcorrespondingauthor{Shen Huang}{huangshenno1@gmail.com}

  \icmlkeywords{Machine Learning, ICML}

  \vskip 0.3in
]



\printAffiliationsAndNotice{}  

\begin{abstract}
Recent advances in persona-centric memory have revealed the powerful capability of multi-agent systems in managing persona memory, especially in conversational scenarios. However, these complex frameworks often suffer from information loss and are fragile across varying scenarios, resulting in suboptimal performance. In this paper, we propose \textbf{\method}, an agentic memory management system that formulates persona-centric memory management as an end-to-end task within a single-agent setting. To further improve the performance of our agentic memory manager, we draw inspiration from the evolution of human memory and synthesize a user-assistant dialogue dataset along with corresponding operation-level memory updating labels. Building on this, we introduce a novel Memory-based Levenshtein Distance to formalize the memory updating reward, and propose a tailored reinforcement learning framework to further enhance the management capabilities of \method{}. Extensive experiments show that both training-free and RL-trained \method{} outperform all product-level baselines across diverse long-term memory benchmarks, including LoCoMo, HaluMem, and PersonaMem.
Code will be open-sourced upon acceptance.
\end{abstract}
\section{Introduction}
In recent years, large language models (LLMs) have showcased tremendous capability across diverse real-life scenarios~\cite{qwen3tech, llmsurvey}. With the increasing of personal usage, mining useful information from User-AI interactions and enhancing LLM's personalised capability has become more and more critical~\cite{persona_survey, zhao2025llms}. Thus, how to build an ancillary long-term persona memory management system becomes a realistic demand~\cite{wildfeedback, zhang2024guided}.

Earlier works such as MemGPT~\cite{memgpt} and MemoryBank~\cite{memorybank} draw inspiration from the design of the operating system and the well-adapted retrieval-augmented generation paradigm, simply consider this task as OS management, and maintain a memory database (\emph{i.e.}, memory bank) through atomised operations including $\mathtt{create}$, $\mathtt{read}$, $\mathtt{update}$, and $\mathtt{delete}$ (CRUD). 

Recent approaches, such as Mem0~\cite{mem0}, advance persona memory management by orchestrating multi-agent systems to maintain the memory bank. These methods typically consist of three sequential steps: \emph{i)} memory extraction, \emph{ii)} memory retrieval, and \emph{iii)} memory updating.
While modular, we argue that enforcing such distinct separation introduces a critical bottleneck: cascading information attenuation, especially in the context of long-term memory. In long-term interaction scenarios, the context is inherently fragmented due to the necessity of session-based slicing to adhere to realistic scenarios~\cite{halumem}. Imposing a multi-agent architecture atop this already-fragmented data exacerbates context loss. In a typical multi-agent setting, the handover between the three aforementioned steps relies on discrete message passing. This process effectively acts as a layer of lossy compression, where subtle contextual dependencies are stripped away.
Prior findings on comparing single-agent and multi-agent~\cite{kim2025sciencescalingagentsystems} have also noticed these limitations.


To bridge this gap, we propose to formulate the persona memory management task as a single-agent setting through an end-to-end manner.
In practice, we employ the well-adapted reasoning-and-act (\emph{i.e.,} ReAct) paradigm to design our memory agent, namely \textbf{\method{}}.
\method{} does not merely process the retrieved memories; it actively reasons over what to retrieve and how to integrate it. This allows the system to dynamically adjust its focus across the entire memory context, mitigating the information loss inherent in static extraction-retrieval pipelines.
Specifically, when ingesting a dialogue session, the memory agent will autonomously incorporate the aforementioned memory extraction and memory retrieval with a search tool to retrieve relevant memories from the existing memory bank. The memory agent is required to output a series of memory operations to update the memory bank and then move to the next session. By iterating the process above, it can maintain the memory bank from scratch in an agentic manner.
Experiments empirically show that \method{} tends to present comparative and even stronger performance compared to production-level multi-agent methods.

To further advance the capabilities of \method{}, we propose transforming the end-to-end setting into a learnable objective. A critical limitation in current research is the lack of datasets that support operation-level evaluation of memory systems. As a result, previous studies have resorted to training the updating module within multi-agent memory systems using downstream task accuracy as a proxy reward in reinforcement learning~\cite{memory-r1}. However, this formulation often leads to severe sparse reward issues, hindering convergence. Inspired by the mechanisms of human memory evolution~\cite{evol-instruct}, we introduce a novel data synthesis methodology that generates user-assistant dialogues along with target memory updating operations. Building on this, we reframe the evaluation metric from an operation-centric to a memory state-centric perspective and propose a tailored RL framework. We formalize the trajectory-level reward by defining a novel memory-based Levenshtein Distance between the generated memory state and the target state. Extensive experimental results across diverse benchmarks and different backbone models show the effectiveness and superiority of this method.

All in all, our core contributions are threefold:
\begin{itemize}
    \item We first formulate the persona memory management task as a single-agent setting through an end-to-end manner, and propose a novel agentic memory system, namely \method{}, for persona-centric scenarios which surpasses all previous multi-agent based methods despite its lightweight design.
    \item We draw inspiration from human memory evolution and design a data synthesis framework for persona memory management, which generates user-assistant dialogues with target memory updating operations.
    \item Grounded on all above, we propose a novel homologous reinforcement learning framework to further improve the performance of \method{}. Our method shows remarkable performance gain on diverse long-term persona memory benchmarks.
\end{itemize}
\section{Related Works}

\subsection{Persona Memory Management for LLMs}

Personalized large language models require mechanisms to store, retrieve, and update user-specific information across long-term interactions~\cite{omem, memos, kang2025memory}.
Early work extends LLMs with external memory modules, treating persona memory as persistent storage.
MemGPT~\cite{memgpt} introduces explicit memory read and write operations inspired by operating systems, while MemoryBank~\cite{memorybank} formalizes persona memory as a structured repository maintained through atomic CRUD operations.
These approaches largely follow a retrieval-augmented generation paradigm and manage memory in a reactive manner, without explicitly modeling its temporal evolution across dialogue sessions, motivating more agentic and learning-based designs.

\subsection{Agentic and Learning-based Memory Systems}

Recent work explores agentic frameworks to improve the autonomy of memory management.
Mem0~\cite{mem0} adopts a multi-agent architecture that decomposes memory management into extraction, retrieval, and updating modules.
LightMem~\cite{lightmem} improves efficiency via a lightweight, human-inspired memory hierarchy that decouples memory construction from online inference.
Beyond training-free designs, Memory-R1~\cite{memory-r1} incorporates reinforcement learning to optimize memory updating within modular pipelines.
In parallel, the ReAct paradigm~\cite{react} demonstrates that LLM-based agents can interleave reasoning with tool use, enabling more autonomous behaviors.
However, existing agentic memory systems remain largely modular and stage-wise, which limits global optimization over long-term memory evolution.
In contrast, we formulate persona memory management as a single-agent, end-to-end decision-making problem optimized with trajectory-level, outcome-based rewards.

\begin{figure*}[ht]
  \begin{center}
    \centerline{\includegraphics[width=\textwidth]{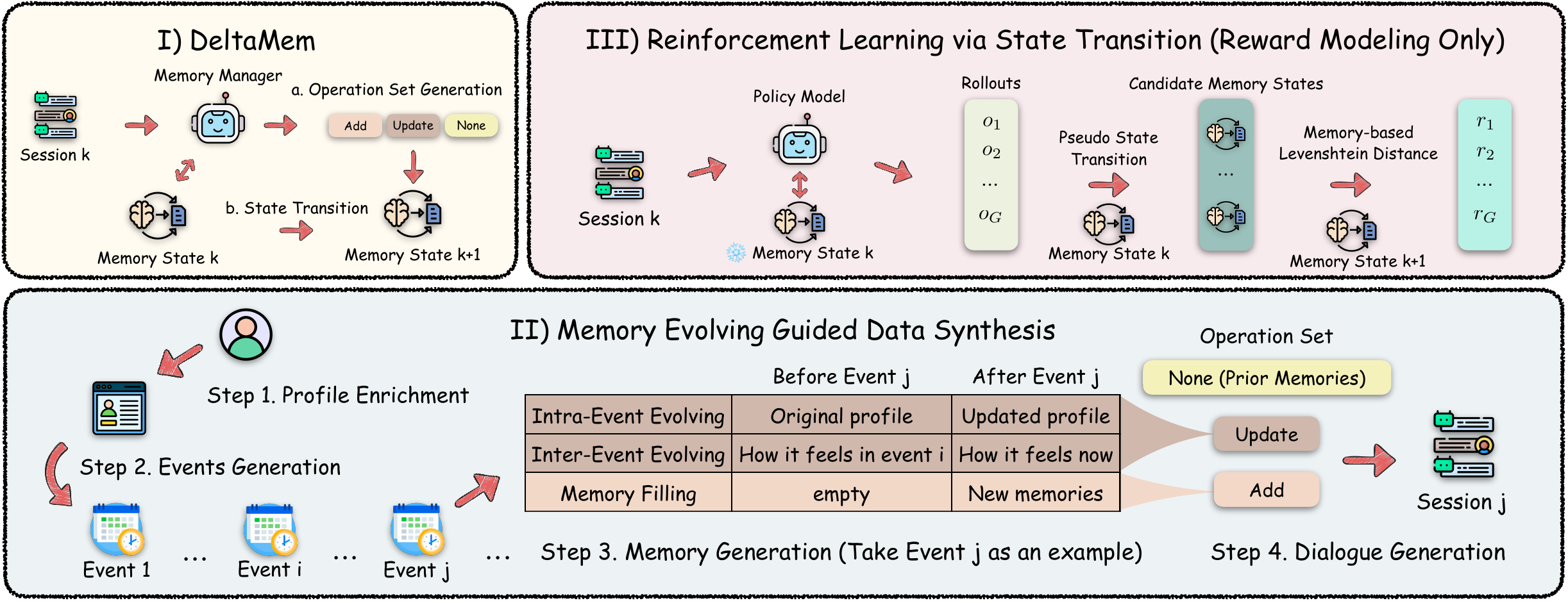}}
    \caption{\emph{i)} Framework of DeltaMem. When ingesting a new session $k$, the memory manager will interact with the current memory state and finally generate a set of operations to update the current memory state. \emph{ii)} Brief illustration of memory evolving guided data synthesis. \emph{iii)} The reward modeling module of reinforcement learning via state transition.}
    \label{fig:main-fig}
  \end{center}
  \vskip -0.25in
\end{figure*}

\section{Methodology}
In this section, we delve into the methodology of our work. Unlike conventional multi-agent memory management systems, we pioneer in building an agentic memory manager, namely \textbf{\method{}}, to manage persona memory in an end-to-end manner.
Then, to further improve the capability of our agentic memory manager, we introduce a novel data synthesis method to synthesize user-assistant dialogues for the memory agent scenarios. Grounded on all the above, we finally propose a tailored reinforcement learning algorithm to enhance the agentic memory manager's memory management capabilities.

\subsection{Overall Framework of \method}
To illustrate the framework of \method, we start by introducing some important concepts.
In this paper, we primarily follow the backend settings of Mem0~\cite{mem0} and introduce a memory bank $\mathbf{m}=\{(m_i,t_i,\mathrm{id}_i)\}_{i=1}^n$ to manage the persona memory across chronological dialogues. Within the memory bank, each memory point is defined as a \textit{factual} and \textit{atomic} sentence $m$ with a corresponding timestamp $t$ and a unique memory ID $\rm id$. Upon this, we start by defining the \textbf{memory state}:
\begin{definition}
    \label{def:memory_state}
    Let $I$ be a set of unique memory IDs, and $\mathcal{M}$ be the universe space of possible memory content-timestamp tuples. Then, a memory state at time $t$ is defined as a finite partial function: $\mathbf{S}_t: I\rightharpoonup \mathcal{M}$. Thus, we build a bijection between the memory bank and the memory state.
\end{definition}

Given $t$ chronological dialogues, an ideal memory management system $f$ should be capable to compress chronological dialogues into a memory state $\mathbf{S}$ without any information loss. Then, for a new session $d_{t+1}$ and the memory state $\mathbf{S}_{t}$ at timing $t$, the ingesting process can be denoted as:
\begin{equation}
    \mathbf{S}_{t+1}=f(\mathbf{S}_t,d_{t+1})
\end{equation}
where $f:\mathbf{S}\times\mathcal{O}\to\mathbf{S}$ is a state transition function that transforms an original state to the next one via an operation space $\mathcal{O}=\{\mathtt{add}, \mathtt{update}\}$. Grounded on this, we formulate $\mathbf{S}_t \to \mathbf{S}_{t+1}$ as the \textbf{memory state transition}.

Based on the aforementioned definitions, we further introduce the framework of \method. As presented in ~\cref{fig:main-fig}, we treat the maintenance of the memory bank as an end-to-end task. When ingesting a brand-new dialogue session, the memory manager is required to output a series of memory operations to update the memory bank before moving to the next session. Specifically, before reaching the final decision, it should reason autonomously to incorporate the aforementioned memory extraction and memory retrieval with a search tool to retrieve relevant memories from the existing memory bank. By iterating the process above, the memory manager can maintain the memory bank across time from scratch in an agentic manner. Notably, training-free DeltaMem surpasses all baselines on diverse benchmarks despite its lightweight design.

\subsection{Memory Evolving Guided Data Synthesis}
However, unlike conventional agentic tasks, memory management training is trapped with a lack of training data, difficulty of direct evaluation, and the multi-agent design. Thus, limited prior research focused on the training of the memory management system. Recent methods like Memory-R1~\cite{memory-r1} attempt to use downstream task performance as the reward signal to train the memory manager with reinforcement learning. We argue that this may introduce an unaffordable sparse reward issue in RL since the answer model is not reliable enough to correctly answer the question. On the other hand, the questions cannot always cover every memory entry in practice.

To address this gap, we draw inspiration from the memory-evolving mechanism of human beings and introduce a novel method to synthesize long-term user-assistant dialogues along with operation-level labels. As illustrated in ~\cref{fig:main-fig}.II, the data synthesis process primarily consists of four sequential steps: 1) \textit{Profile Enrichment}: synthesize a \textbf{detailed profile} and \textbf{initial memory state} based on a seed persona, 2) \textit{Event Generation}: synthesize \textbf{chronological event summaries} based on the user profile, 3) \textit{Memory Generation}: for each event, synthesize \textbf{diverse memory operations and their corresponding keywords}, and 4) \textit{Dialogue Generation}: for each event, synthesize \textbf{user-assistant dialogues} based on prior information. \cref{alg:data_syn} presents a brief illustration of the data synthesis process. 

Considering that memory generation serves as the core of the data synthesis framework, we primarily introduce three major components of memory generation and leave the detailed illustration in ~\cref{app:data_synthesis}.

\begin{algorithm}[h]
  \caption{Memory Evolving Guided Data Synthesis}
  \label{alg:data_syn}
  \begin{algorithmic}
    \STATE {\bfseries Input:} $\rm persona\_seed$, data synthesizer $\rm LLM$
    \STATE Enrich basic profile and initialize memory state $\mathbf{S}_0$: $\mathbb{P},\mathbf{S}_0={\rm LLM(persona\_seed)}$
    \STATE Generate $n$ chronological events: $\{\mathbb{E}_i\}_{i=1}^n={\rm LLM}(\mathbb{P})$
    \FOR{$i=1$ {\bfseries to} $n$}
    \STATE Generate memory updating operation set: $o_i=\{(m,{\rm op},\mathcal{K})\} = {\rm LLM}(\mathbb{P},\mathbb{E}_i)$
    \STATE Generate dialogues: $d_i={\rm LLM}(\mathbb{P},\mathbb{E}_i,o_i)$
    \ENDFOR
    \STATE {\bfseries Output:} basic profile $\mathbb{P}$, initial memory state $\mathbf{S}_0$, dialogues and corresponding memory updating operation set $\{d_i,o_i\}_{i=1}^n$
  \end{algorithmic}
\end{algorithm}

\paragraph{Intra-Event Memory Evolving} We first address intra-event memory evolution, defined as the process by which specific events trigger alterations in user characteristics. We categorize these alterations into four domains: career, health, relationship, and preference. Furthermore, we define four distinct evolution mechanisms: 1) expand, 2) adjust, 3) shift, and 4) partial deletion. The model is tasked with selecting the relevant attributes from the user's existing profile and determining the appropriate evolution mechanism to apply based on the context of the current event.

\paragraph{Inter-Event Memory Evolving} Subsequently, we address inter-event memory evolution, modeling how subsequent events retrospectively alter a user's perception of preceding ones. We first establish links between semantically related events and designate the chronologically earliest instance as the source of a stable \textit{ memory anchor}. To update this anchor, we introduce three distinct mechanisms of evolution: emotional updates, meaning updates, and shifts in self-belief. The model is tasked with selecting the most appropriate mechanism to update the memory iteratively, thereby integrating new context into the original anchor.

\paragraph{Memory Filling} Finally, leveraging the event summaries and the previously constructed memory update operations, we synthesize the remaining memory addition operations. To enhance robustness, a small fraction of semantically relevant memories from preceding events are reused as the content for $\mathtt{none}$ operations. Thus, we successfully establish the complete set of memory operations for a dialogue session, encompassing $\mathtt{add}$, $\mathtt{update}$, and $\mathtt{none}$.

In summary, we introduce a novel method to synthesize chronological dialogue and operation set pairs in long-term scenarios. Specifically, for any session $d_{t}$ at timing $t$, our method provide its prior memory state $\mathbf{S}_t$ and corresponding operation set $o_t=\{(m,{\rm op},\mathcal{K})\}$, and thereby get the next memory state $\mathbf{S}_{t+1}$ inherently via conducting memory state transition with $\mathbf{S}_t$ and $o$.

\subsection{Reinforcement Learning via State Transition}
Beyond synthesizing end-to-end training data, long-term memory management still struggles with proper training algorithms.
\citeauthor{memory-r1} primarily consider the performance of downstream tasks as the reward signal for reinforcement learning. We argue that these methods are limited since the memory agents do not directly associate with these tasks, but with a shared memory bank instead. On the other hand, these tasks cannot always cover every memory entry within a session. Thus, direct evaluation of memory updating should be proposed under memory agent scenarios.

In this section, we dive into the homologous reinforcement learning framework for \method{}. We first illustrate the motivation behind state-level evaluation, and then move to our proposed memory-based Levenshtein distance, which serves as the core design for the reward function used in reinforcement learning. Finally, we present the overall formulation of reinforcement learning of \method{}.

\subsubsection{Motivation: From Operation to State}
Given a dialogue session and corresponding memory updating operations, a naive method is to verify the generated operations with the oracle operations (\emph{i.e.}, direct evaluation over memory updating). However, this practicality is limited since it's difficult to propose a proper metric for cross-type operation cases. Thus, to avoid the difficulty of direct operation-level evaluation, we consider this problem from the perspective of the memory state the generated operations lead to, instead of the operations themselves.

Then, the question turns to be quantifying the semantic transition cost required to transform the predicted memory state $\mathbf{S}_{pred}$ into the target state $\mathbf{S}_{target}$. We formulate this transformation process by analyzing the divergence between the two states. Specifically, we isolate the distinct components of each state using relative complements:
\begin{equation}
    \Delta_{pred} = \mathbf{S}_{pred} \setminus \mathbf{S}_{target}, \quad \Delta_{target} = \mathbf{S}_{target} \setminus \mathbf{S}_{pred}
\end{equation}

Here, $\Delta_{pred}$ represents the \textit{possibly redundant} memory entries for the prediction (candidates for deletion or substitution), while $\Delta_{target}$ represents the \textit{possibly missing} entries for the target state (targets for insertion or substitution). Thus, the core idea is to distinguish the soft information intersection between $\Delta_{pred}$ and $\Delta_{target}$.

\subsubsection{Memory-based Levenshtein Distance}
\label{sec:memory-based-ld}
To further quantify the semantic transition cost between $\Delta_{pred}$ and $\Delta_{target}$, we propose a metric termed the Memory-based Levenshtein Distance.
In a strict symbolic matching regime, the distance would simply be $|\Delta_{pred}| + |\Delta_{target}|$, treating all divergence as pure insertion and deletion errors. However, in the context of natural language, an entry in $\Delta_{pred}$ may be semantically equivalent to an entry in $\Delta_{target}$ despite lexical differences. Such a pair should be treated as a semantic substitution rather than independent insertion and deletion operations. To capture this, we model the alignment between $\Delta_{pred}$ and $\Delta_{target}$ as an Optimal Transport problem. Specifically, we compute a semantic similarity matrix $\mathbf{\Phi} \in \mathbb{R}^{|\Delta_{pred}| \times |\Delta_{target}|}$, where each element $\Phi_{ij}$ quantifies the semantic affinity between the $i$-th element of $\Delta_{pred}$ and the $j$-th element of $\Delta_{target}$. The core of our metric involves finding an optimal matching strategy $\gamma^*$ that maximizes the total semantic validity of the substitutions:
\begin{equation}
    \gamma^* = \mathop{\arg\max}_{\gamma \in \Gamma} \sum_{(i,j) \in \gamma} \Phi_{ij}
\end{equation}
where $\Gamma$ represents the set of all valid one-to-one matching between the two relative complement sets. To avoid low-similarity matching, we introduce a similarity threshold $\tau$ to filter those pairs whose similarity is lower than $\tau$ in $\Gamma$. Besides, we also remove matchings source from unexpected $\mathtt{update}$ to avoid reward hacking. This optimal alignment $\gamma^*$ effectively identifies the \textit{substitution} operations, separating semantically correct variations from true discrepancies, thereby providing a robust measure of the state transition difficulty. While the embedding-based alignment $\gamma^*$ captures high-level semantic correspondence, embedding space density can sometimes lead to false positives. For example, one candidate memory entry is semantically related to a target one but missing some factual details. To mitigate this, we incorporate a keyword coverage metric, namely \textbf{local lexical fidelity}, as a fine-grained verification step.

For each target memory entry $y_j \in \Delta_{target}$, we use the pre-defined set of keywords $\mathcal{K}_j = \{k_1, k_2, \dots, k_M\}$ as the factual anchors for that memory. For an aligned pair consisting of a predicted entry $x_i \in \Delta_{pred}$ and a target entry $y_j \in \Delta_{target}$, the local fidelity score $s_{ij}$ is defined as the proportion of these ground truth keywords that are present in the predicted sentence:
\begin{equation}
    s_{ij} = \frac{\sum_{k \in \mathcal{K}_j} \mathbb{I}(k \in x_i)}{|\mathcal{K}_j| + \epsilon}
\end{equation}
where $\mathbb{I}(\cdot)$ is the indicator function which equals 1 if the keyword $k$ appears in the text of $x_i$, and 0 otherwise. This score $s_{ij} \in [0, 1]$ ensures that the model is rewarded not merely for vector-space proximity, but for explicitly recalling specific, pre-determined factual details.

By aggregating the keyword coverage scores, we finalize to formulate the memory-based Levenshtein Distance:
\begin{equation}
    \begin{split}
        \mathrm{dist}(\Delta_{pred}, \Delta_{target})^+&=|\Delta_{pred}|-\sum_{(i,j) \in \gamma^*} s_{ij}\\
        \mathrm{dist}(\Delta_{pred}, \Delta_{target})^-&=|\Delta_{target}|-\sum_{(i,j) \in \gamma^*} s_{ij}
    \end{split}
\end{equation}
Where the positive distance indicates the counts of false positives in $\Delta_{pred}$ while the negative distance indicates the counts of true negatives in $\Delta_{target}$.

\subsubsection{State Transition-Based Reward Modeling}
To compute the effective soft counts of true positives, we further normalize the memory-based Levenshtein Distance against the total volume of the transition sets to derive soft precision $P_{soft}$ and soft recall $R_{soft}$:
\begin{equation}
    P_{soft} = \frac{\sum_{(i,j) \in \gamma^*} s_{ij}}{|\Delta_{pred}| + \epsilon}, R_{soft} = \frac{\sum_{(i,j) \in \gamma^*} s_{ij}}{|\Delta_{target}| + \epsilon}
\end{equation}

Soft precision measures the validity of the generated updates. It penalizes the model for generating entries in $\Delta_{pred}$ that fail to align significantly with any entry in $\Delta_{target}$. On the other hand, soft recall measures the completeness of the memory recovery. It penalizes the model for failing to cover entries in $\Delta_{target}$.

To provide a balanced optimization objective that discourages both cautious under-generation (high precision, low recall) and aggressive over-generation (high recall, low precision), we utilize the soft F1 score as the final reward $r_{trans}$ for state transition:
\begin{equation}
    r_{trans} = 2 \cdot \frac{P_{soft} \cdot R_{soft}}{P_{soft} + R_{soft} + \epsilon}
\end{equation}
This reward acts as a proxy for the inverse of the memory-based Levenshtein Distance, which inherently aligns with the aforementioned soft information intersection. By maximizing $r_{trans}$, the policy $\pi_\theta$ is encouraged to minimize the edit costs, thereby navigating the memory state towards the target standard.

\begin{table*}[]
\centering
\caption{Performance of different methods on LoCoMo. We use GPT-4o-mini as the answer model for all methods. As for the memory agent, all training-free baselines are evaluated with GPT-4o-mini. Note that Memory-R1 trains both the answer model and one module of its multi-agent system, and these two models are \textbf{directly trained} on LoCoMo. B1, F1, and LJ stand for BLEU-1 score, F1 score and LLM-as-a-Judge score, respectively.}
\label{tab:locomo}
\resizebox{\textwidth}{!}{%
\begin{tabular}{@{}cccccccccccccccc@{}}
\toprule
\multicolumn{1}{c|}{} &
  \multicolumn{3}{c|}{Single Hop} &
  \multicolumn{3}{c|}{Multi Hop} &
  \multicolumn{3}{c|}{Open Domain} &
  \multicolumn{3}{c|}{Temperal} &
  \multicolumn{3}{c}{Overall} \\
\multicolumn{1}{c|}{\multirow{-2}{*}{Method}} &
  BS &
  F1 &
  \multicolumn{1}{c|}{LJ} &
  BS &
  F1 &
  \multicolumn{1}{c|}{LJ} &
  BS &
  F1 &
  \multicolumn{1}{c|}{LJ} &
  BS &
  F1 &
  \multicolumn{1}{c|}{LJ} &
  BS &
  F1 &
  LJ \\ \midrule
\multicolumn{16}{c}{\textit{\textbf{Training-free   Multi-Agent Memory Manager}}} \\ \midrule
\multicolumn{1}{c|}{A-Mem} &
  27.58 &
  33.34 &
  \multicolumn{1}{c|}{54.05} &
  14.90 &
  20.76 &
  \multicolumn{1}{c|}{39.79} &
  8.81 &
  9.22 &
  \multicolumn{1}{c|}{18.85} &
  31.08 &
  35.40 &
  \multicolumn{1}{c|}{49.91} &
  24.82 &
  29.96 &
  48.38 \\
\multicolumn{1}{c|}{Mem0} &
  35.34 &
  43.17 &
  \multicolumn{1}{c|}{60.88} &
  22.27 &
  33.99 &
  \multicolumn{1}{c|}{52.84} &
  18.19 &
  23.34 &
  \multicolumn{1}{c|}{40.62} &
  44.53 &
  54.21 &
  \multicolumn{1}{c|}{59.50} &
  33.79 &
  42.55 &
  57.86 \\
\multicolumn{1}{c|}{LangMem} &
  33.63 &
  40.91 &
  \multicolumn{1}{c|}{71.12} &
  26.86 &
  35.51 &
  \multicolumn{1}{c|}{62.23} &
  {\ul 22.32} &
  26.04 &
  \multicolumn{1}{c|}{47.92} &
  25.84 &
  30.75 &
  \multicolumn{1}{c|}{23.43} &
  30.06 &
  36.88 &
  58.11 \\
\multicolumn{1}{c|}{Zep} &
  38.92 &
  49.56 &
  \multicolumn{1}{c|}{76.60} &
  23.30 &
  35.74 &
  \multicolumn{1}{c|}{61.70} &
  14.82 &
  19.37 &
  \multicolumn{1}{c|}{41.35} &
  34.53 &
  42.00 &
  \multicolumn{1}{c|}{49.31} &
  33.64 &
  43.57 &
  65.99 \\
\multicolumn{1}{c|}{LightMem} &
  37.59 &
  47.55 &
  \multicolumn{1}{c|}{72.06} &
  22.46 &
  33.06 &
  \multicolumn{1}{c|}{53.90} &
  19.38 &
  24.77 &
  \multicolumn{1}{c|}{44.79} &
  39.93 &
  53.94 &
  \multicolumn{1}{c|}{69.16} &
  34.17 &
  44.81 &
  66.43 \\ \midrule
\multicolumn{16}{c}{\textit{\textbf{RL-trained   Multi-Agent Memory Manager}}} \\ \midrule
\multicolumn{1}{c|}{{\color[HTML]{9B9B9B} Memory-R1-7B}} &
  {\color[HTML]{9B9B9B} 37.59} &
  {\color[HTML]{9B9B9B} 47.55} &
  \multicolumn{1}{c|}{{\color[HTML]{9B9B9B} 67.81}} &
  {\color[HTML]{9B9B9B} 38.49} &
  {\color[HTML]{9B9B9B} 47.75} &
  \multicolumn{1}{c|}{{\color[HTML]{9B9B9B} 49.61}} &
  {\color[HTML]{9B9B9B} 19.38} &
  {\color[HTML]{9B9B9B} 24.77} &
  \multicolumn{1}{c|}{{\color[HTML]{9B9B9B} 20.71}} &
  {\color[HTML]{9B9B9B} 23.55} &
  {\color[HTML]{9B9B9B} 40.96} &
  \multicolumn{1}{c|}{{\color[HTML]{9B9B9B} 69.16}} &
  {\color[HTML]{9B9B9B} 26.06} &
  {\color[HTML]{9B9B9B} 33.64} &
  {\color[HTML]{9B9B9B} 62.34} \\ \midrule
\multicolumn{16}{c}{\textit{\textbf{Training-free   Agentic Memory Manager}}} \\ \midrule
\multicolumn{1}{c|}{DeltaMem-4B} &
  40.87 &
  49.79 &
  \multicolumn{1}{c|}{75.62} &
  25.87 &
  37.72 &
  \multicolumn{1}{c|}{61.35} &
  19.81 &
  25.52 &
  \multicolumn{1}{c|}{47.92} &
  45.26 &
  56.73 &
  \multicolumn{1}{c|}{70.09} &
  37.73 &
  47.51 &
  70.13 \\
\multicolumn{1}{c|}{DeltaMem-8B} &
  38.58 &
  48.71 &
  \multicolumn{1}{c|}{77.41} &
  \textbf{27.63} &
  \textbf{38.62} &
  \multicolumn{1}{c|}{61.70} &
  19.45 &
  25.41 &
  \multicolumn{1}{c|}{46.88} &
  44.54 &
  57.66 &
  \multicolumn{1}{c|}{69.78} &
  36.63 &
  47.27 &
  71.04 \\
\multicolumn{1}{c|}{DeltaMem-4o-mini} &
  42.51 &
  53.75 &
  \multicolumn{1}{c|}{{\ul 80.38}} &
  {\ul 27.15} &
  {\ul 38.18} &
  \multicolumn{1}{c|}{58.87} &
  20.28 &
  {\ul 26.59} &
  \multicolumn{1}{c|}{47.92} &
  \textbf{49.70} &
  {\ul 59.10} &
  \multicolumn{1}{c|}{{\ul 73.21}} &
  \textbf{39.81} &
  {\ul 50.32} &
  72.92 \\ \midrule
\multicolumn{16}{c}{\textit{\textbf{RL-trained Agentic   Memory Manager}}} \\ \midrule
\multicolumn{1}{c|}{DeltaMem-4B-RL} &
  {\ul 43.05} &
  {\ul 53.96} &
  \multicolumn{1}{c|}{80.02} &
  26.71 &
  38.09 &
  \multicolumn{1}{c|}{\textbf{65.60}} &
  19.97 &
  26.46 &
  \multicolumn{1}{c|}{\textbf{51.04}} &
  46.29 &
  57.75 &
  \multicolumn{1}{c|}{72.59} &
  39.29 &
  50.13 &
  {\ul 74.03} \\
\multicolumn{1}{c|}{DeltaMem-8B-RL} &
  \textbf{43.40} &
  \textbf{54.57} &
  \multicolumn{1}{c|}{\textbf{82.05}} &
  25.90 &
  37.44 &
  \multicolumn{1}{c|}{{\ul 64.18}} &
  \textbf{22.67} &
  \textbf{29.04} &
  \multicolumn{1}{c|}{{\ul \textbf{51.04}}} &
  {\ul 47.61} &
  \textbf{58.79} &
  \multicolumn{1}{c|}{\textbf{73.83}} &
  {\ul 39.78} &
  \textbf{50.72} &
  \textbf{75.13} \\ \bottomrule
\end{tabular}%
}
\end{table*}

\subsubsection{Reinforcement Learning Formulation}
Apart from the state transition-based reward, we additionally introduce two binary rewards, \emph{i.e.}, format reward and retrieval reward. Specifically, we use a format reward to check whether the trajectory follows the ReAct paradigm. Retrieval reward is used to check whether the policy model can recall the specific existing memories in operation $\mathtt{update}$ and $\mathtt{add}$. The overall reward is formulated as:
\begin{equation}
    r = 0.1\cdot r_{format} + 0.1\cdot r_{retrieval} + 0.8\cdot r_{trans}
\end{equation}

Finally, we follow the standard GRPO~\cite{grpo} to design our training framework as detailed in~\cref{app:grpo}.
During RL training, input $x$ is a user-assistant dialogue session at timing $t$, and the policy model consistently interact with current memory state $\mathbf{S}_t$. After generating the final operations, $r_{trans}$ is calculated according to $\mathbf{S}_{t+1}$ and the generated pseudo memory states as illustrated in ~\cref{fig:main-fig} for further RL training.

\section{Experiments}
\subsection{Experimental Setup}
\paragraph{Training Details} For RL training, we use the widely adopted RL framework \texttt{VeRL}\footnote{\url{https://github.com/volcengine/verl}}. Besides, for foundation models, we incorporate Qwen3-4B and Qwen3-8B~\cite{qwen3tech} as our base models.

\paragraph{Benchmarks} In this paper, we evaluate our methods on three commonly-used benchmarks, including LoCoMo~\cite{locomo}, HaluMem~\cite{halumem}, and PersonaMem~\cite{personamem}.

\paragraph{Baselines} We include extensive training-free and training-based baselines to present the superiority of our methods.
For training-free multi-agent pipelines, we consider methods including A-Mem~\cite{amem}, Mem0~\cite{mem0}, LangMem, Zep~\cite{zep}, Memobase, Supermemory, Mirix~\cite{mirix}, and LightMem~\cite{lightmem}.
For training-based methods, since there is limited work focusing on this, we only choose Memory-R1~\cite{memory-r1} as our baseline.

We present more detailed experimental settings in~\cref{app:data_synthesis}.

\subsection{Main Results}
\paragraph{LoCoMo} Table~\ref{tab:locomo} presents the comprehensive evaluation on the LoCoMo benchmark, where our proposed \method{} framework consistently outperforms both training-free and RL-trained baselines across all query categories. Most notably, the RL-optimized variant, \method{}-8B-RL, achieves a state-of-the-art Overall LLM-as-a-Judge score of 75.13 and an F1 score of 50.72, establishing a significant margin over the strongest baseline, LightMem, and the proprietary Zep system. The effectiveness of our reinforcement learning stage is evidenced by the clear performance jump between the training-free and RL-trained agentic managers; for instance, RL fine-tuning yields a 4.1 point increase in LJ score for the 8B model, demonstrating that optimizing the memory policy is critical for handling complex, dynamic contexts. Furthermore, our method exhibits exceptional robustness in reasoning-intensive tasks, particularly in the Temporal category, where it scores 73.83, effectively solving the challenge of memory evolution that limits static retrieval methods like Mem0.

\begin{table*}[h]
\centering
\caption{Performance of different methods on PersonaMem. Pref-Rec, Recall-Reason, New-Ideas, Pref-Evol, Recall-User, Gen-New, and Recall-Facts stand for providing preference-aligned recommendations, recalling the reasons behind previous updates, suggesting new ideas, tracking full preference evolution, recalling user-shared facts, generalizing to new scenarios, and recalling facts mentioned by the user, respectively. Following common settings, each question was evaluated by 3 separate runs.}
\label{tab:personamem}
\resizebox{\textwidth}{!}{%
\begin{tabular}{@{}ccccccccc@{}}
\toprule
\multicolumn{1}{c|}{Methods} &
  Pref-Rec &
  Recall-Reason &
  New-Ideas &
  Pref-Evol &
  Recall-User &
  Gen-New &
  \multicolumn{1}{c|}{Recall-Facts} &
  Overall \\ \midrule
\multicolumn{9}{c}{\textit{\textbf{Training-free   Multi-Agent Memory Manager}}}                                            \\ \midrule
\multicolumn{1}{c|}{Mirix}       & 30.91 & 63.64 & 17.20       & 48.92 & 34.11 & 19.30 & \multicolumn{1}{c|}{41.18} & 38.37 \\
\multicolumn{1}{c|}{Mem0}        & 46.67 & 77.10 & 19.35       & 45.32 & 36.95 & 28.65 & \multicolumn{1}{c|}{41.18} & 43.12 \\ 
\multicolumn{1}{c|}{LightMem}    & 37.58 & 82.15 & 14.70       & 50.36 & 48.58 & 14.62 & \multicolumn{1}{c|}{43.14} & 44.82 \\
\multicolumn{1}{c|}{Supermemory} & 54.55 & 79.12 & 26.88       & 54.44 & 53.65 & 52.30 & \multicolumn{1}{c|}{54.90} & 53.88 \\
\multicolumn{1}{c|}{Zep}         & 60.61 & 80.47 & {\ul 36.92} & 52.52 & 57.88 & 51.46 & \multicolumn{1}{c|}{56.86} & 56.71 \\
\multicolumn{1}{c|}{Memobase}    & 63.64 & 82.31 & 32.62       & 56.52 & 64.71 & 53.22 & \multicolumn{1}{c|}{62.27} & 58.89 \\ \midrule
\multicolumn{9}{c}{\textit{\textbf{Training-free   Agentic Memory Manager}}}                                                \\ \midrule
\multicolumn{1}{c|}{DeltaMem-4B} & 64.24 & 82.83 & 20.43       & 50.84 & 57.36 & 59.65 & \multicolumn{1}{c|}{70.59} & 55.52 \\
\multicolumn{1}{c|}{DeltaMem-8B} &
  65.45 &
  82.49 &
  24.73 &
  52.52 &
  56.07 &
  \textbf{69.01} &
  \multicolumn{1}{c|}{64.71} &
  57.10 \\
\multicolumn{1}{c|}{DeltaMem-4o-mini} &
  58.79 &
  {\ul 83.16} &
  29.75 &
  54.68 &
  \textbf{66.93} &
  59.65 &
  \multicolumn{1}{c|}{\textbf{76.47}} &
  59.71 \\ \midrule
\multicolumn{9}{c}{\textit{\textbf{RL-trained Agentic   Memory Manager}}}                                                   \\ \midrule
\multicolumn{1}{c|}{DeltaMem-4B-RL} &
  \textbf{69.70} &
  82.49 &
  34.77 &
  {\ul 56.35} &
  57.11 &
  64.91 &
  \multicolumn{1}{c|}{74.51} &
  {\ul 60.10} \\
\multicolumn{1}{c|}{DeltaMem-8B-RL} &
  {\ul 67.27} &
  \textbf{85.19} &
  \textbf{40.14} &
  \textbf{58.27} &
  {\ul 65.12} &
  {\ul 66.67} &
  \multicolumn{1}{c|}{\textbf{76.47}} &
  \textbf{63.61} \\ \bottomrule
\end{tabular}%
}
\vskip -0.1in
\end{table*}
\begin{table}[h]
\centering
\caption{Performance of different methods on HaluMem.}
\resizebox{0.9\columnwidth}{!}{%
\begin{tabular}{@{}cccc@{}}
\toprule
\multicolumn{1}{c|}{Method} & \multicolumn{1}{c|}{\makecell{Memory\\Extraction}} & \multicolumn{1}{c|}{\makecell{Memory\\Updating}} & \makecell{Question\\Answering} \\ \midrule
\multicolumn{4}{c}{\textit{\textbf{Training-free Multi-Agent Memory Manager}}}                              \\ \midrule
\multicolumn{1}{c|}{Zep}           & \multicolumn{1}{c|}{-}     & \multicolumn{1}{c|}{\textbf{47.28}} & 55.47 \\
\multicolumn{1}{c|}{Mem0}          & \multicolumn{1}{c|}{57.31} & \multicolumn{1}{c|}{25.50} & 53.02 \\
\multicolumn{1}{c|}{Mem0-Graph}    & \multicolumn{1}{c|}{57.85} & \multicolumn{1}{c|}{24.50} & 54.66 \\
\multicolumn{1}{c|}{Memobase}      & \multicolumn{1}{c|}{25.13} & \multicolumn{1}{c|}{5.20}  & 35.33 \\
\multicolumn{1}{c|}{Supermemory}   & \multicolumn{1}{c|}{56.90} & \multicolumn{1}{c|}{16.37} & 54.07 \\
\multicolumn{1}{c|}{LightMem}      & \multicolumn{1}{c|}{70.80} & \multicolumn{1}{c|}{7.43}  & 58.38 \\ \midrule
\multicolumn{4}{c}{\textit{\textbf{Training-free Agentic Memory Manager}}}                           \\ \midrule
\multicolumn{1}{c|}{DeltaMem-4B}    & \multicolumn{1}{c|}{65.06} & \multicolumn{1}{c|}{25.40} & 60.14 \\
\multicolumn{1}{c|}{DeltaMem-8B}    & \multicolumn{1}{c|}{68.02} & \multicolumn{1}{c|}{31.77} & 62.50 \\ 
\multicolumn{1}{c|}{DeltaMem-4o}    & \multicolumn{1}{c|}{73.13} & \multicolumn{1}{c|}{35.11} & 60.89 \\ \midrule
\multicolumn{4}{c}{\textit{\textbf{RL-trained Agentic Memory Manager}}}                              \\ \midrule
\multicolumn{1}{c|}{DeltaMem-4B-RL} & \multicolumn{1}{c|}{{\ul 79.57}} & \multicolumn{1}{c|}{38.89} & {\ul 65.62} \\
\multicolumn{1}{c|}{DeltaMem-8B-RL} & \multicolumn{1}{c|}{\textbf{80.65}} & \multicolumn{1}{c|}{{\ul 41.54}} & \textbf{66.43} \\ \bottomrule
\end{tabular}%
}
\label{tab:halumem}
\end{table}
\paragraph{HaluMem}
Table~\ref{tab:halumem} details the performance on the HaluMem dataset, evaluating the capabilities of memory extraction, updating, and downstream question answering. Following a general setting, all methods are evaluated with GPT-4o as the backend model of memory management except \method-4/8B series. \method-8B-RL demonstrates superior performance in the final \textit{Question Answering} task, achieving a score of \textbf{66.43}, significantly surpassing both the best training-free baseline, LightMem, and the commercial solution Zep. This dominance is driven by substantial gains in memory extraction; our RL optimization boosts the extraction score of the 8B model from 68.02 (training-free) to \textbf{80.65}, verifying that the agent learns to identify salient information more effectively. While Zep achieves the highest score in \textit{Memory Updating}, it fails to translate this into superior downstream performance. In contrast, \method-8B-RL maintains a highly competitive updating score while securing the best overall QA performance. This indicates that our method offers a more balanced and effective pipeline, avoiding the pitfalls of baselines like LightMem, which suffers from critically low updating capabilities, and ensuring that extracted and updated memories are actually useful for the final response.

\paragraph{PersonaMem}
Table~\ref{tab:personamem} details the performance on the PersonaMem benchmark, where DeltaMem-8B-RL achieves a state-of-the-art Overall score of 63.61, surpassing the strongest training-free baseline, Memobase, and significantly outperforming widely used systems like Zep and Mem0. The impact of Reinforcement Learning is most pronounced in generative tasks; for instance, in the New-Ideas metric, DeltaMem-8B-RL scores 40.14 compared to 24.73 for its training-free counterpart, representing a 62\% relative improvement in synthesizing novel suggestions. Furthermore, our method balances this generative capability with rigorous precision, as DeltaMem-8B-RL achieves a top-tier score of 76.47 in Recall-Facts, proving that the agent can evolve user preferences without hallucinating or losing core historical details.

\subsection{Training Dynamics}
\begin{figure}[h]
  \begin{center}
    \centerline{\includegraphics[width=\columnwidth]{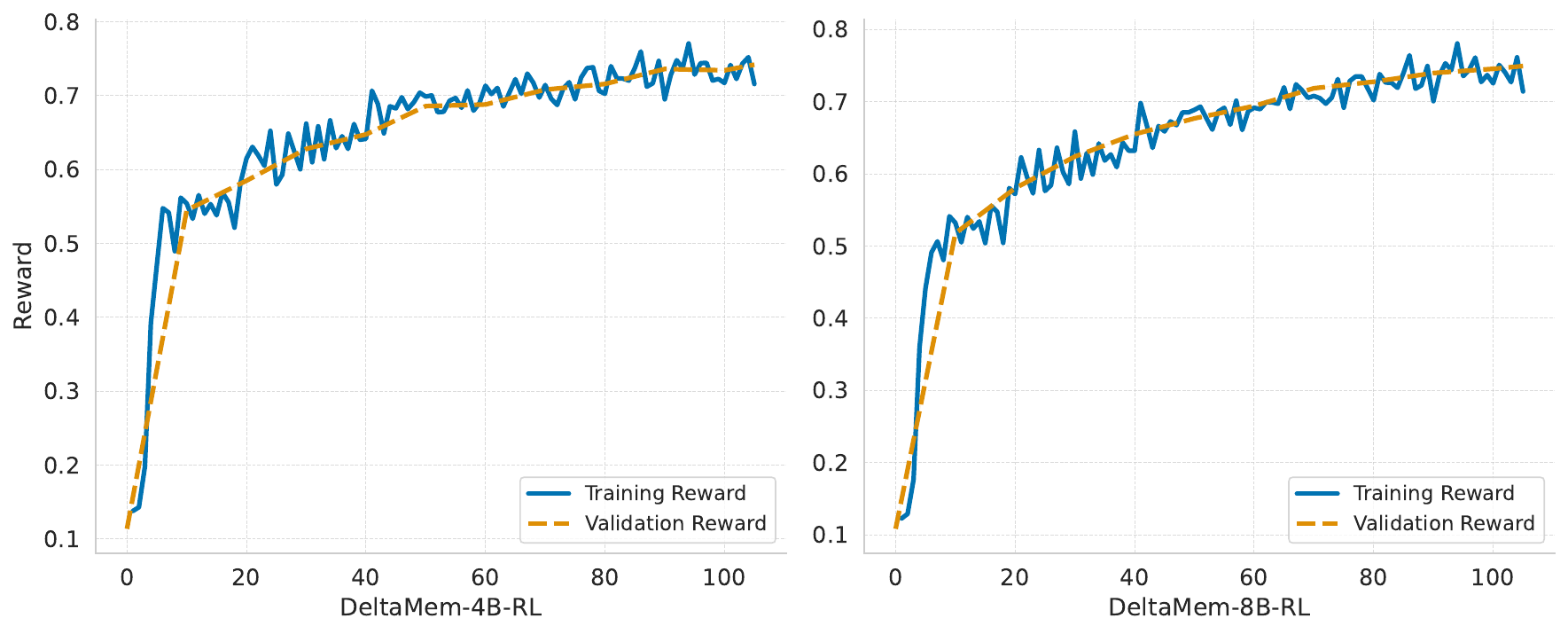}}
    \caption{Reward dynamics during RL training of \method{}-4B-RL (left) and \method{}-8B-RL (right).}
    \label{fig:reward}
  \end{center}
  \vskip -0.2in
\end{figure}

We further analyze the training dynamics of the RL stage in Figure~\ref{fig:reward}. Both the DeltaMem-4B-RL and DeltaMem-8B-RL models exhibit a rapid initial performance gain followed by a steady convergence, with rewards stabilizing at approximately 0.75. Crucially, the validation reward tracks the training reward closely throughout the optimization process; this strong alignment indicates that our proposed reward facilitates robust generalization without overfitting, ensuring the agent learns a consistent policy for memory management.

\section{Ablation and Analysis}
\subsection{Effectiveness of Local Lexical Fidelity}
\begin{figure}[ht]
  \begin{center}
    \centerline{\includegraphics[width=0.9\columnwidth]{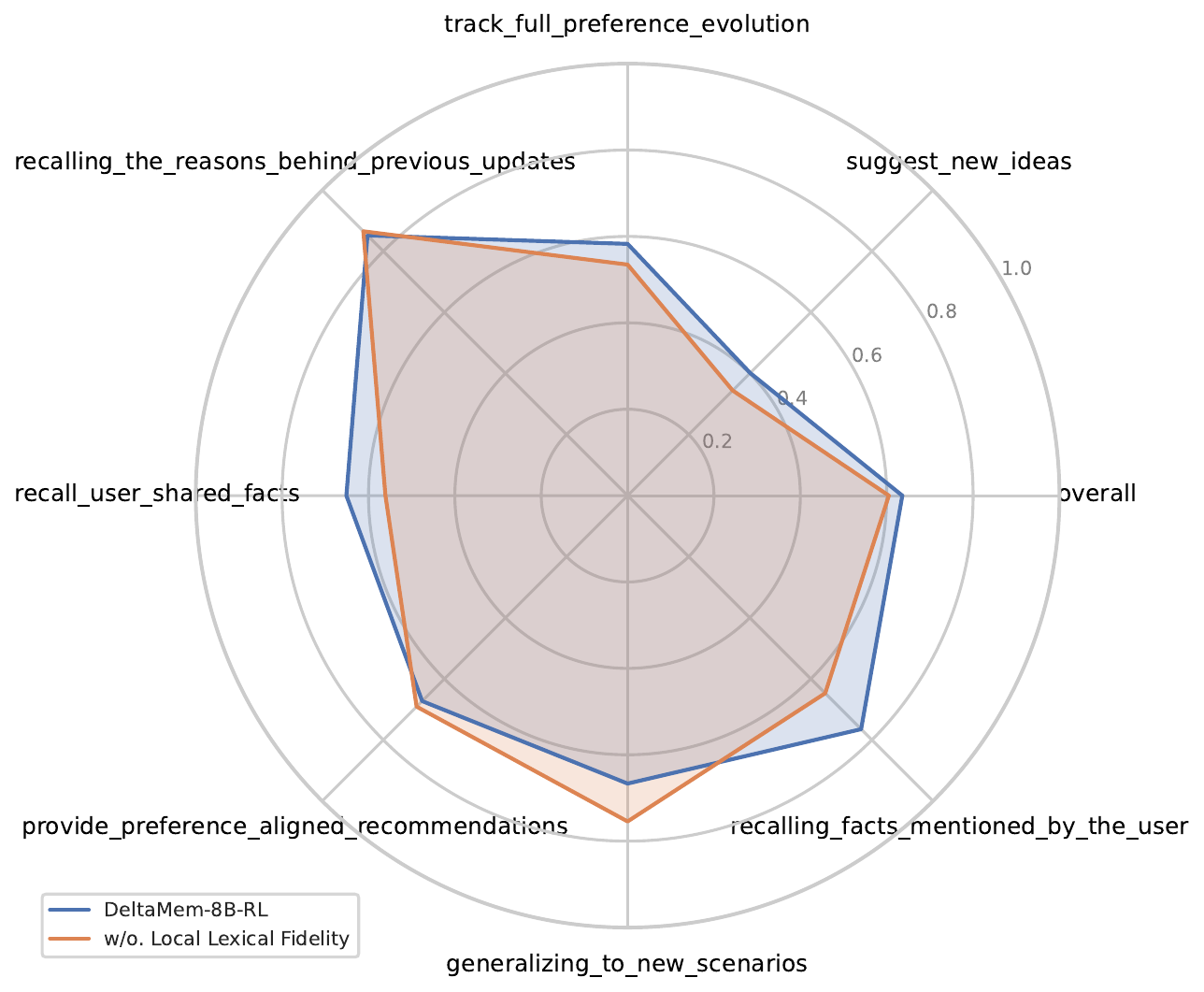}}
    \caption{Ablation study on local lexical fidelity.}
    \label{fig:llf}
  \end{center}
  \vskip -0.2in
\end{figure}

To present the effectiveness of local lexical fidelity, we replace the local fidelity score in memory-based Levenshtein Distance with its original similarity and keep all other parameters to train Qwen3-8B. As shown in Figure~\ref{fig:llf}, the original DeltaMem-8b-RL shows a remarkable performance gain on recalling user-shared facts and recalling facts mentioned by the user, which implies that the local lexical fidelity can boost the memory manager's fact extraction capability and preserve the integrity of personal memory. Specifically, the original DeltaMem-8b-RL lags behind in generalizing to new scenarios, which is attributed to a similar reason that more detailed information might lead to overconfidence on downstream tasks. Overall, DeltaMem-8b-RL got around 3 points of performance gain on the PersonaMem benchmark compared to the model trained without local lexical fidelity, which shows the effectiveness of this design.

\subsection{Impact of Threshold $\tau$}
In Section~\ref{sec:memory-based-ld}, we introduce a hyperparameter $\tau$ to filter the less similar matchings before calculating memory-based Levenshtein Distance. To further analyze the impact of the threshold $\tau$, we conduct ablation experiments by changing it from 0.0 to 0.9. 
\begin{table}[ht]
\centering
\caption{Impact of threshold $\tau$ on LoCoMo.}
\label{tab:threshold}
\resizebox{0.9\columnwidth}{!}{%
\begin{tabular}{@{}c|ccccccc@{}}
\toprule
Threshold & 0.00  & 0.50  & 0.60  & 0.70  & 0.75           & 0.80           & 0.90  \\ \midrule
B1        & 38.40 & 37.12 & 38.35 & 39.24 & \textbf{39.87} & {\ul 39.74}    & 38.32 \\
F1        & 48.92 & 47.93 & 48.75 & 50.37 & {\ul 50.72}    & \textbf{50.78} & 48.75 \\ \bottomrule
\end{tabular}%
}
\vskip -0.1in
\end{table}

As shown in Table~\ref{tab:threshold}, when $\tau$ is set to a smaller value, the desired matchings will result in wrong reward signals and lead to a performance drop on LoCoMo. Moreover, an extremely large $\tau$ (\emph{i.e.}, 0.9) will also ruin the performance of our method. Notably, performance is quite stable when $\tau$ is set to range from 0.7 to 0.8. Thus, in this paper, we set $\tau$ to 0.75 for all models.

\subsection{Stable Performance across Time}
\begin{figure}[ht]
  \begin{center}
    \centerline{\includegraphics[width=0.9
    \columnwidth]{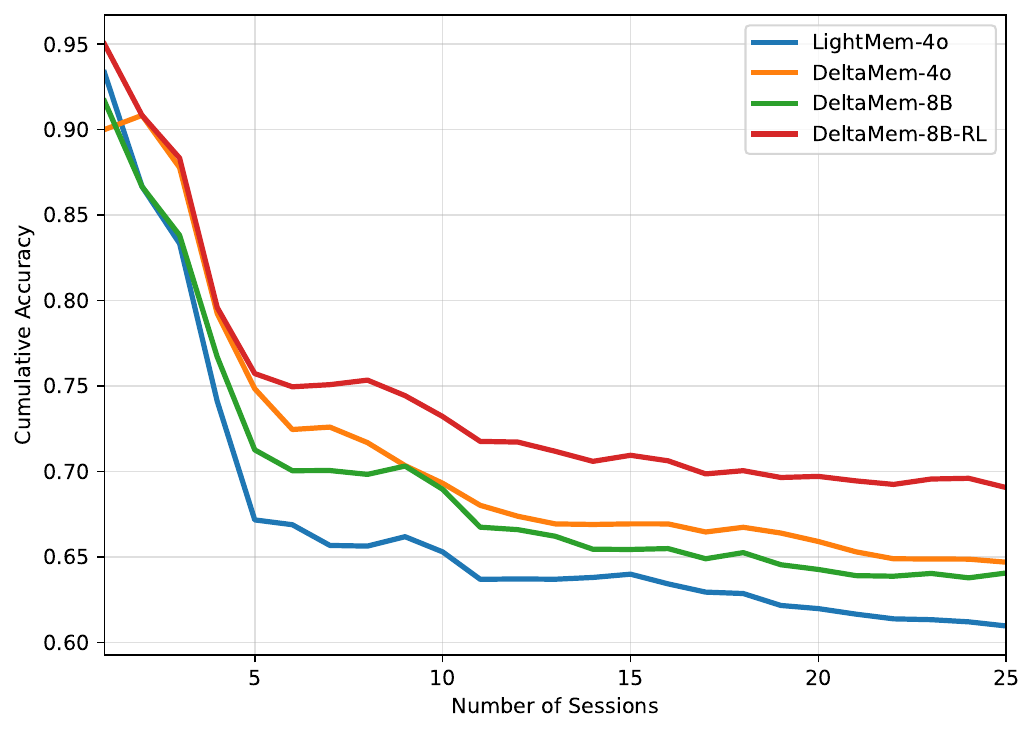}}
    \caption{Average cumulative accuracy on HaluMem as the number of sessions increases.}
    \label{fig:acc}
  \end{center}
  \vskip -0.4in
\end{figure}

Since HaluMem provides question answering after every session, we further present the performance dynamics of our method across time. Specifically, we evaluate the average cumulative accuracy of the first 25 sessions for 20 users in HaluMem. As shown in Figure~\ref{fig:acc}, all methods present best performance when starting long-term memory management but consistently drop as the number of sessions increases. \method{} consistently surpasses LightMem as the memory bank expands, demonstrating the superiority of our proposed \method{}. Besides, the accuracy curve of \method{} tends to be more stable across time.

\section{Conclusion}
In this paper, we point out the common issues in the widely-adapted multi-agent memory management systems. To address these issues, we first propose \textbf{\method{}}, which formulates the memory management as an end-to-end task. Despite its lightweight design, training-free \method{} got comparative, even better performance on all the benchmarks. Upon this, we draw inspiration from the state transition of human memory, and propose a novel data synthesis method along with a homologous RL algorithm to further improve the capability of \method{}. Experiments show the superiority of our method across diverse dimensions including memory extraction, memory updating and memory-augmented question answering.

\nocite{langley00}

\newpage
\section*{Impact Statement}
This paper presents \method{}, an end-to-end framework for long-term persona memory management. To further boost the performance, we propose a method for synthesizing user-AI conversations using personas derived from PersonaHub and a homologous RL framework. Our work contributes to the development of robust agentic memory systems by providing scalable, diverse training and evaluation data without the privacy concerns associated with scraping real user logs.

However, we acknowledge several potential risks. First, while we employ random sampling from PersonaHub to maximize diversity, synthetic personas generated by LLMs may still exhibit or amplify social stereotypes present in the underlying model's training data. Consequently, the synthesized conversations should be viewed as an approximation of human behavior, not a perfect substitute. Second, the widespread generation of synthetic text poses a risk of data contamination for future models. To mitigate this, we release our dataset with clear metadata indicating its synthetic nature. Finally, researchers should exercise caution when optimizing solely on synthetic metrics, as they may diverge from real-world user preferences.
\bibliography{main}

@article{grpo,
  author       = {Zhihong Shao and
                  Peiyi Wang and
                  Qihao Zhu and
                  Runxin Xu and
                  Junxiao Song and
                  Mingchuan Zhang and
                  Y. K. Li and
                  Y. Wu and
                  Daya Guo},
  title        = {DeepSeekMath: Pushing the Limits of Mathematical Reasoning in Open
                  Language Models},
  journal      = {CoRR},
  volume       = {abs/2402.03300},
  year         = {2024},
  url          = {https://doi.org/10.48550/arXiv.2402.03300},
  doi          = {10.48550/ARXIV.2402.03300},
  eprinttype    = {arXiv},
  eprint       = {2402.03300},
  bibsource    = {dblp computer science bibliography, https://dblp.org}
}

@misc{kim2025sciencescalingagentsystems,
      title={Towards a Science of Scaling Agent Systems}, 
      author={Yubin Kim and Ken Gu and Chanwoo Park and Chunjong Park and Samuel Schmidgall and A. Ali Heydari and Yao Yan and Zhihan Zhang and Yuchen Zhuang and Mark Malhotra and Paul Pu Liang and Hae Won Park and Yuzhe Yang and Xuhai Xu and Yilun Du and Shwetak Patel and Tim Althoff and Daniel McDuff and Xin Liu},
      year={2025},
      eprint={2512.08296},
      archivePrefix={arXiv},
      primaryClass={cs.AI},
      url={https://arxiv.org/abs/2512.08296}, 
}

@article{memgpt,
  author       = {Charles Packer and
                  Vivian Fang and
                  Shishir G. Patil and
                  Kevin Lin and
                  Sarah Wooders and
                  Joseph E. Gonzalez},
  title        = {MemGPT: Towards LLMs as Operating Systems},
  journal      = {CoRR},
  volume       = {abs/2310.08560},
  year         = {2023},
  url          = {https://doi.org/10.48550/arXiv.2310.08560},
  doi          = {10.48550/ARXIV.2310.08560},
  eprinttype    = {arXiv},
  eprint       = {2310.08560},
  bibsource    = {dblp computer science bibliography, https://dblp.org}
}

@inproceedings{memorybank,
  author       = {Wanjun Zhong and
                  Lianghong Guo and
                  Qiqi Gao and
                  He Ye and
                  Yanlin Wang},
  editor       = {Michael J. Wooldridge and
                  Jennifer G. Dy and
                  Sriraam Natarajan},
  title        = {MemoryBank: Enhancing Large Language Models with Long-Term Memory},
  booktitle    = {Thirty-Eighth {AAAI} Conference on Artificial Intelligence, {AAAI}
                  2024, Thirty-Sixth Conference on Innovative Applications of Artificial
                  Intelligence, {IAAI} 2024, Fourteenth Symposium on Educational Advances
                  in Artificial Intelligence, {EAAI} 2014, February 20-27, 2024, Vancouver,
                  Canada},
  pages        = {19724--19731},
  publisher    = {{AAAI} Press},
  year         = {2024},
  url          = {https://doi.org/10.1609/aaai.v38i17.29946},
  doi          = {10.1609/AAAI.V38I17.29946},
  bibsource    = {dblp computer science bibliography, https://dblp.org}
}

@article{mem0,
  author       = {Prateek Chhikara and
                  Dev Khant and
                  Saket Aryan and
                  Taranjeet Singh and
                  Deshraj Yadav},
  title        = {Mem0: Building Production-Ready {AI} Agents with Scalable Long-Term
                  Memory},
  journal      = {CoRR},
  volume       = {abs/2504.19413},
  year         = {2025},
  url          = {https://doi.org/10.48550/arXiv.2504.19413},
  doi          = {10.48550/ARXIV.2504.19413},
  eprinttype    = {arXiv},
  eprint       = {2504.19413},
  bibsource    = {dblp computer science bibliography, https://dblp.org}
}

@article{memory-r1,
  author       = {Sikuan Yan and
                  Xiufeng Yang and
                  Zuchao Huang and
                  Ercong Nie and
                  Zifeng Ding and
                  Zonggen Li and
                  Xiaowen Ma and
                  Hinrich Sch{\"{u}}tze and
                  Volker Tresp and
                  Yunpu Ma},
  title        = {Memory-R1: Enhancing Large Language Model Agents to Manage and Utilize
                  Memories via Reinforcement Learning},
  journal      = {CoRR},
  volume       = {abs/2508.19828},
  year         = {2025},
  url          = {https://doi.org/10.48550/arXiv.2508.19828},
  doi          = {10.48550/ARXIV.2508.19828},
  eprinttype    = {arXiv},
  eprint       = {2508.19828},
  bibsource    = {dblp computer science bibliography, https://dblp.org}
}

@inproceedings{react,
  author       = {Shunyu Yao and
                  Jeffrey Zhao and
                  Dian Yu and
                  Nan Du and
                  Izhak Shafran and
                  Karthik R. Narasimhan and
                  Yuan Cao},
  title        = {ReAct: Synergizing Reasoning and Acting in Language Models},
  booktitle    = {The Eleventh International Conference on Learning Representations,
                  {ICLR} 2023, Kigali, Rwanda, May 1-5, 2023},
  publisher    = {OpenReview.net},
  year         = {2023},
  url          = {https://openreview.net/forum?id=WE\_vluYUL-X},
  bibsource    = {dblp computer science bibliography, https://dblp.org}
}

@article{omem,
  author       = {Piaohong Wang and
                  Motong Tian and
                  Jiaxian Li and
                  Yuan Liang and
                  Yuqing Wang and
                  Qianben Chen and
                  Tiannan Wang and
                  Zhicong Lu and
                  Jiawei Ma and
                  Yuchen Eleanor Jiang and
                  Wangchunshu Zhou},
  title        = {O-Mem: Omni Memory System for Personalized, Long Horizon, Self-Evolving
                  Agents},
  journal      = {CoRR},
  volume       = {abs/2511.13593},
  year         = {2025},
  url          = {https://doi.org/10.48550/arXiv.2511.13593},
  doi          = {10.48550/ARXIV.2511.13593},
  eprinttype    = {arXiv},
  eprint       = {2511.13593},
  bibsource    = {dblp computer science bibliography, https://dblp.org}
}

@article{memos,
  author       = {Zhiyu Li and
                  Shichao Song and
                  Chenyang Xi and
                  Hanyu Wang and
                  Chen Tang and
                  Simin Niu and
                  Ding Chen and
                  Jiawei Yang and
                  Chunyu Li and
                  Qingchen Yu and
                  Jihao Zhao and
                  Yezhaohui Wang and
                  Peng Liu and
                  Zehao Lin and
                  Pengyuan Wang and
                  Jiahao Huo and
                  Tianyi Chen and
                  Kai Chen and
                  Kehang Li and
                  Zhen Tao and
                  Junpeng Ren and
                  Huayi Lai and
                  Hao Wu and
                  Bo Tang and
                  Zhenren Wang and
                  Zhaoxin Fan and
                  Ningyu Zhang and
                  Linfeng Zhang and
                  Junchi Yan and
                  Mingchuan Yang and
                  Tong Xu and
                  Wei Xu and
                  Huajun Chen and
                  Haofeng Wang and
                  Hongkang Yang and
                  Wentao Zhang and
                  Zhi{-}Qin John Xu and
                  Siheng Chen and
                  Feiyu Xiong},
  title        = {MemOS: {A} Memory {OS} for {AI} System},
  journal      = {CoRR},
  volume       = {abs/2507.03724},
  year         = {2025},
  url          = {https://doi.org/10.48550/arXiv.2507.03724},
  doi          = {10.48550/ARXIV.2507.03724},
  eprinttype    = {arXiv},
  eprint       = {2507.03724},
  bibsource    = {dblp computer science bibliography, https://dblp.org}
}

@article{lightmem,
  author       = {Jizhan Fang and
                  Xinle Deng and
                  Haoming Xu and
                  Ziyan Jiang and
                  Yuqi Tang and
                  Ziwen Xu and
                  Shumin Deng and
                  Yunzhi Yao and
                  Mengru Wang and
                  Shuofei Qiao and
                  Huajun Chen and
                  Ningyu Zhang},
  title        = {LightMem: Lightweight and Efficient Memory-Augmented Generation},
  journal      = {CoRR},
  volume       = {abs/2510.18866},
  year         = {2025},
  url          = {https://doi.org/10.48550/arXiv.2510.18866},
  doi          = {10.48550/ARXIV.2510.18866},
  eprinttype    = {arXiv},
  eprint       = {2510.18866},
  bibsource    = {dblp computer science bibliography, https://dblp.org}
}

@article{qwen3tech,
  author       = {An Yang and
                  Anfeng Li and
                  Baosong Yang and
                  Beichen Zhang and
                  Binyuan Hui and
                  Bo Zheng and
                  Bowen Yu and
                  Chang Gao and
                  Chengen Huang and
                  Chenxu Lv and
                  Chujie Zheng and
                  Dayiheng Liu and
                  Fan Zhou and
                  Fei Huang and
                  Feng Hu and
                  Hao Ge and
                  Haoran Wei and
                  Huan Lin and
                  Jialong Tang and
                  Jian Yang and
                  Jianhong Tu and
                  Jianwei Zhang and
                  Jian Yang and
                  Jiaxi Yang and
                  Jingren Zhou and
                  Junyang Lin and
                  Kai Dang and
                  Keqin Bao and
                  Kexin Yang and
                  Le Yu and
                  Lianghao Deng and
                  Mei Li and
                  Mingfeng Xue and
                  Mingze Li and
                  Pei Zhang and
                  Peng Wang and
                  Qin Zhu and
                  Rui Men and
                  Ruize Gao and
                  Shixuan Liu and
                  Shuang Luo and
                  Tianhao Li and
                  Tianyi Tang and
                  Wenbiao Yin and
                  Xingzhang Ren and
                  Xinyu Wang and
                  Xinyu Zhang and
                  Xuancheng Ren and
                  Yang Fan and
                  Yang Su and
                  Yichang Zhang and
                  Yinger Zhang and
                  Yu Wan and
                  Yuqiong Liu and
                  Zekun Wang and
                  Zeyu Cui and
                  Zhenru Zhang and
                  Zhipeng Zhou and
                  Zihan Qiu},
  title        = {Qwen3 Technical Report},
  journal      = {CoRR},
  volume       = {abs/2505.09388},
  year         = {2025},
  url          = {https://doi.org/10.48550/arXiv.2505.09388},
  doi          = {10.48550/ARXIV.2505.09388},
  eprinttype    = {arXiv},
  eprint       = {2505.09388},
  bibsource    = {dblp computer science bibliography, https://dblp.org}
}

@inproceedings{locomo,
  author       = {Adyasha Maharana and
                  Dong{-}Ho Lee and
                  Sergey Tulyakov and
                  Mohit Bansal and
                  Francesco Barbieri and
                  Yuwei Fang},
  editor       = {Lun{-}Wei Ku and
                  Andre Martins and
                  Vivek Srikumar},
  title        = {Evaluating Very Long-Term Conversational Memory of {LLM} Agents},
  booktitle    = {Proceedings of the 62nd Annual Meeting of the Association for Computational
                  Linguistics (Volume 1: Long Papers), {ACL} 2024, Bangkok, Thailand,
                  August 11-16, 2024},
  pages        = {13851--13870},
  publisher    = {Association for Computational Linguistics},
  year         = {2024},
  url          = {https://doi.org/10.18653/v1/2024.acl-long.747},
  doi          = {10.18653/V1/2024.ACL-LONG.747},
  bibsource    = {dblp computer science bibliography, https://dblp.org}
}

@article{halumem,
  author       = {Ding Chen and
                  Simin Niu and
                  Kehang Li and
                  Peng Liu and
                  Xiangping Zheng and
                  Bo Tang and
                  Xinchi Li and
                  Feiyu Xiong and
                  Zhiyu Li},
  title        = {HaluMem: Evaluating Hallucinations in Memory Systems of Agents},
  journal      = {CoRR},
  volume       = {abs/2511.03506},
  year         = {2025},
  url          = {https://doi.org/10.48550/arXiv.2511.03506},
  doi          = {10.48550/ARXIV.2511.03506},
  eprinttype    = {arXiv},
  eprint       = {2511.03506},
  bibsource    = {dblp computer science bibliography, https://dblp.org}
}

@article{personamem,
  author       = {Bowen Jiang and
                  Zhuoqun Hao and
                  Young{-}Min Cho and
                  Bryan Li and
                  Yuan Yuan and
                  Sihao Chen and
                  Lyle H. Ungar and
                  Camillo J. Taylor and
                  Dan Roth},
  title        = {Know Me, Respond to Me: Benchmarking LLMs for Dynamic User Profiling
                  and Personalized Responses at Scale},
  journal      = {CoRR},
  volume       = {abs/2504.14225},
  year         = {2025},
  url          = {https://doi.org/10.48550/arXiv.2504.14225},
  doi          = {10.48550/ARXIV.2504.14225},
  eprinttype    = {arXiv},
  eprint       = {2504.14225},
  bibsource    = {dblp computer science bibliography, https://dblp.org}
}

@article{persona_survey,
  author       = {Zhehao Zhang and
                  Ryan A. Rossi and
                  Branislav Kveton and
                  Yijia Shao and
                  Diyi Yang and
                  Hamed Zamani and
                  Franck Dernoncourt and
                  Joe Barrow and
                  Tong Yu and
                  Sungchul Kim and
                  Ruiyi Zhang and
                  Jiuxiang Gu and
                  Tyler Derr and
                  Hongjie Chen and
                  Junda Wu and
                  Xiang Chen and
                  Zichao Wang and
                  Subrata Mitra and
                  Nedim Lipka and
                  Nesreen K. Ahmed and
                  Yu Wang},
  title        = {Personalization of Large Language Models: {A} Survey},
  journal      = {Trans. Mach. Learn. Res.},
  volume       = {2025},
  year         = {2025},
  url          = {https://openreview.net/forum?id=tf6A9EYMo6},
  bibsource    = {dblp computer science bibliography, https://dblp.org}
}

@article{amem,
  author       = {Wujiang Xu and
                  Zujie Liang and
                  Kai Mei and
                  Hang Gao and
                  Juntao Tan and
                  Yongfeng Zhang},
  title        = {{A-MEM:} Agentic Memory for {LLM} Agents},
  journal      = {CoRR},
  volume       = {abs/2502.12110},
  year         = {2025},
  url          = {https://doi.org/10.48550/arXiv.2502.12110},
  doi          = {10.48550/ARXIV.2502.12110},
  eprinttype    = {arXiv},
  eprint       = {2502.12110},
  bibsource    = {dblp computer science bibliography, https://dblp.org}
}

@article{zep,
  author       = {Preston Rasmussen and
                  Pavlo Paliychuk and
                  Travis Beauvais and
                  Jack Ryan and
                  Daniel Chalef},
  title        = {Zep: {A} Temporal Knowledge Graph Architecture for Agent Memory},
  journal      = {CoRR},
  volume       = {abs/2501.13956},
  year         = {2025},
  url          = {https://doi.org/10.48550/arXiv.2501.13956},
  doi          = {10.48550/ARXIV.2501.13956},
  eprinttype    = {arXiv},
  eprint       = {2501.13956},
  bibsource    = {dblp computer science bibliography, https://dblp.org}
}

@article{wildfeedback,
  author       = {Taiwei Shi and
                  Zhuoer Wang and
                  Longqi Yang and
                  Ying{-}Chun Lin and
                  Zexue He and
                  Mengting Wan and
                  Pei Zhou and
                  Sujay Kumar Jauhar and
                  Xiaofeng Xu and
                  Xia Song and
                  Jennifer Neville},
  title        = {WildFeedback: Aligning LLMs With In-situ User Interactions And Feedback},
  journal      = {CoRR},
  volume       = {abs/2408.15549},
  year         = {2024},
  url          = {https://doi.org/10.48550/arXiv.2408.15549},
  doi          = {10.48550/ARXIV.2408.15549},
  eprinttype    = {arXiv},
  eprint       = {2408.15549},
  bibsource    = {dblp computer science bibliography, https://dblp.org}
}

@article{mirix,
  author       = {Yu Wang and
                  Xi Chen},
  title        = {{MIRIX:} Multi-Agent Memory System for LLM-Based Agents},
  journal      = {CoRR},
  volume       = {abs/2507.07957},
  year         = {2025},
  url          = {https://doi.org/10.48550/arXiv.2507.07957},
  doi          = {10.48550/ARXIV.2507.07957},
  eprinttype    = {arXiv},
  eprint       = {2507.07957},
  bibsource    = {dblp computer science bibliography, https://dblp.org}
}

@inproceedings{zhao2025llms,
  author       = {Siyan Zhao and
                  Mingyi Hong and
                  Yang Liu and
                  Devamanyu Hazarika and
                  Kaixiang Lin},
  title        = {Do LLMs Recognize Your Preferences? Evaluating Personalized Preference
                  Following in LLMs},
  booktitle    = {The Thirteenth International Conference on Learning Representations,
                  {ICLR} 2025, Singapore, April 24-28, 2025},
  publisher    = {OpenReview.net},
  year         = {2025},
  url          = {https://openreview.net/forum?id=QWunLKbBGF},
  bibsource    = {dblp computer science bibliography, https://dblp.org}
}

@article{zhang2024guided,
  author       = {Jiarui Zhang},
  title        = {Guided Profile Generation Improves Personalization with LLMs},
  journal      = {CoRR},
  volume       = {abs/2409.13093},
  year         = {2024},
  url          = {https://doi.org/10.48550/arXiv.2409.13093},
  doi          = {10.48550/ARXIV.2409.13093},
  eprinttype    = {arXiv},
  eprint       = {2409.13093},
  bibsource    = {dblp computer science bibliography, https://dblp.org}
}

@article{llmsurvey,
  author       = {Andrea Matarazzo and
                  Riccardo Torlone},
  title        = {A Survey on Large Language Models with some Insights on their Capabilities
                  and Limitations},
  journal      = {CoRR},
  volume       = {abs/2501.04040},
  year         = {2025},
  url          = {https://doi.org/10.48550/arXiv.2501.04040},
  doi          = {10.48550/ARXIV.2501.04040},
  eprinttype    = {arXiv},
  eprint       = {2501.04040},
  bibsource    = {dblp computer science bibliography, https://dblp.org}
}

@inproceedings{evol-instruct,
  author       = {Can Xu and
                  Qingfeng Sun and
                  Kai Zheng and
                  Xiubo Geng and
                  Pu Zhao and
                  Jiazhan Feng and
                  Chongyang Tao and
                  Qingwei Lin and
                  Daxin Jiang},
  title        = {WizardLM: Empowering Large Pre-Trained Language Models to Follow Complex
                  Instructions},
  booktitle    = {The Twelfth International Conference on Learning Representations,
                  {ICLR} 2024, Vienna, Austria, May 7-11, 2024},
  publisher    = {OpenReview.net},
  year         = {2024},
  url          = {https://openreview.net/forum?id=CfXh93NDgH},
  bibsource    = {dblp computer science bibliography, https://dblp.org}
}

@article{kang2025memory,
  author       = {Jiazheng Kang and
                  Mingming Ji and
                  Zhe Zhao and
                  Ting Bai},
  title        = {Memory {OS} of {AI} Agent},
  journal      = {CoRR},
  volume       = {abs/2506.06326},
  year         = {2025},
  url          = {https://doi.org/10.48550/arXiv.2506.06326},
  doi          = {10.48550/ARXIV.2506.06326},
  eprinttype    = {arXiv},
  eprint       = {2506.06326},
  bibsource    = {dblp computer science bibliography, https://dblp.org}
}

@article{zhang2025qwen3embeddingadvancingtext,
  author       = {Yanzhao Zhang and
                  Mingxin Li and
                  Dingkun Long and
                  Xin Zhang and
                  Huan Lin and
                  Baosong Yang and
                  Pengjun Xie and
                  An Yang and
                  Dayiheng Liu and
                  Junyang Lin and
                  Fei Huang and
                  Jingren Zhou},
  title        = {Qwen3 Embedding: Advancing Text Embedding and Reranking Through Foundation
                  Models},
  journal      = {CoRR},
  volume       = {abs/2506.05176},
  year         = {2025},
  url          = {https://doi.org/10.48550/arXiv.2506.05176},
  doi          = {10.48550/ARXIV.2506.05176},
  eprinttype    = {arXiv},
  eprint       = {2506.05176},
  bibsource    = {dblp computer science bibliography, https://dblp.org}
}
\bibliographystyle{icml2026}

\newpage
\appendix
\onecolumn
\section{Memory Evolving Guided Data Synthesis}
\label{app:data_synthesis}
As presented in Figure~\ref{fig:main-fig}.II, the overall data synthesis process consists of four steps: \emph{i)} profile enrichment, \emph{ii)} event generation, \emph{iii)} memory generation, and \emph{iv)} dialogue generation. In this section, we will illustrate each step in detail. The detailed prompts for each step are presented in the Supplement Materials.

\paragraph{Profile Enrichment} We primarily conduct profile enrichment based on the profile generation step of HaluMem.

\paragraph{Event Generation}
To generate longitudinally consistent user trajectories, we task the language model with synthesizing a chronological sequence of $N=25$ discrete events $E = \{\mathbb{E}_1, \dots, \mathbb{E}_n\}$ spanning a 1--3 year horizon. The generation is conditioned on two distinct input priors: a set of immutable traits $\mathcal{P}_{\text{static}}$ (denoted as \texttt{BASIC\_PROFILE}) and an evolving memory bank $\mathcal{P}_{\text{dyn}}$ (denoted as \texttt{DYNAMIC\_FACTS}). We enforce distributional constraints across $D=6$ semantic domains (e.g., work, health, finance) to ensure holistic lifestyle coverage. Crucially, to mitigate the hallucination of conflicting facts over long contexts, we implement a \textit{state-delta tracking} mechanism. For each event $e_t$, the model must explicitly output a state update vector $\Delta S_t$ targeting four specific tracks: career, health, relationships, and preferences. These updates are governed by a strictly defined set of operators $\mathcal{O} = \{\texttt{new}, \texttt{expand}, \texttt{adjust}, \texttt{shift}, \texttt{partial\_deletion}\}$. We enforce an ``information preservation invariant'' where any modification to an existing fact (operations in $\mathcal{O} \setminus \{\texttt{new}\}$) requires the model to reproduce the exact string of the prior state before generating the update, thereby ensuring referential integrity and causal continuity across the temporal sequence.

\paragraph{Memory Generation}
To simulate realistic cognitive retention, we employ a dual-strategy approach targeting both the longitudinal evolution of narrative threads and the granular extraction of atomic facts. 

First, to simulate the dynamic plasticity of human cognition, we introduce the Inter-Event Memory Evolving protocol. Unlike static knowledge retrieval, this module treats memory as a malleable state subject to iterative refinement based on temporal progression. Given a synthetic user profile $P$ and a chronological sequence of events $\{\mathbb{E}_0, \dots, \mathbb{E}_n\}$, the model isolates a single "memory anchor" originating in $\mathbb{E}_0$. For each subsequent event $\mathbb{E}_{i>0}$, the system generates an updated memory representation $M_i$ that is constrained to strictly subsume prior details while applying a specific cognitive transformation: emotional update (affective shift), meaning update (semantic reinterpretation), or self-belief (identity adjustment). This constraint forces the model to move beyond append-only logging, ensuring that $M_i$ reflects how new experiences recontextualize existing facts. The output is structured as a chronological JSON object, providing a granular trace of how a specific factual thread evolves, coherently and cumulatively, across a user’s interaction history.

Second, to ensure a comprehensive and contiguous representation of the user's state, we employ a Memory Filling module designed to synthesize granular, non-redundant factual memories. By conditioning an LLM on the \texttt{USER\_PROFILE}, chronological \texttt{PAST\_EVENTS}, and the specific \texttt{CURRENT\_EVENT}, this step identifies narrative gaps left by \texttt{EXISTING\_MEMORIES}. The prompt enforces strict syntactic constraints, requiring the generation of 6–10 standalone, third-person declarative statements (approx. 15 words each) that explicitly capture specific entities, future intentions, and implied attitudes without contradicting prior context. Furthermore, the model is tasked with extracting structured keyword triplets and a dialogue summary, outputting a valid JSON object that integrates seamlessly into the synthetic user's long-term memory store.

\paragraph{Dialogue Generation}
To construct a dataset of high-fidelity, legally grounded user-agent interactions, we employed a comprehensive prompt engineering framework designed to enforce strict factual adherence and temporal consistency. As illustrated in~\cref{fig:main-fig}, the synthesis process treats dialogue generation as a constrained constraint-satisfaction problem. The prompt receives a structured context, comprising a synthetic user profile, a timeline of past and current events, and a specific set of knowledge triples (facts to be added, updated, or recalled) and mandates a multi-turn conversation that integrates these elements organically. To mitigate hallucination and ensure causal logic, we enforced a "user-first" information flow constraint, where the AI assistant is prohibited from referencing specific entities or events until they are explicitly introduced by the user. Furthermore, the prompt instantiates an internal Chain-of-Thought process via a mandatory "Work Plan," requiring the model to pre-calculate a "fact introduction schedule" to ensure information is distributed evenly across $N_{min}$ to $N_{max}$ turns rather than clustered. The final output is structured as a parseable JSON object containing both the dialogue history and a metadata layer mapping specific fact IDs to the exact turn indices where they were introduced, facilitating automated validation of coverage and consistency.

\newpage
\section{GRPO Formulation}
\label{app:grpo}
Given a policy model $\pi_{\theta_\text{old}}$ and a reference model $\pi_{\theta_\text{ref}}$, based on $G$ rollouts $\{y_i\}_{i=1}^G\sim\pi_{\theta_\text{old}}(\cdot|x)$ for input $x\sim \mathcal{D}$, the objective of \method{} is to optimize the policy $\pi_{\theta}$ by maximizing the following objective:
\begin{equation}
    \begin{split}
        \mathcal{J}(\theta)=\mathbb{E}_{x\sim D,\{y_i\}_{i=1}^G\sim\pi_{\theta_\text{old}}(\cdot|x)}\frac{1}{G}\sum_{i=1}^{G}\frac{1}{|y_i|}\sum_{t=1}^{|y_i|}\bigg[\min\Big(\frac{\pi_{\theta}(y_{i,t}|x,y_{i,<t})}{\pi_\text{old}(y_{i,t}|x,y_{i,<t})}{A}_{i},\\
        \text{clip}\Big(\frac{\pi_{\theta}(y_{i,t}|x,y_{i,<t})}{\pi_\text{old}(y_{i,t}|x,y_{i,<t})},1-\epsilon,1+\epsilon\Big){A}_{i}\Big)-\beta\mathbb{D}_\text{KL}(\pi_{\theta}|\pi_\text{ref})\bigg]
    \end{split}
\label{eq:grpo}
\end{equation}, where $A_i=\frac{r_i-\text{mean}\left(\left\{r_j\right\}_{j=1}^G\right)}{\text{std}\left(\left\{r_j\right\}_{j=1}^G\right)}$ is the group normalized advantage, and $\epsilon$ is the clipping threshold. The KL divergence term regularizes the policy to remain close to the reference model, weighted by the coefficient $\beta$.
\section{Experimental Setup}
\label{app:setup}
\subsection{Data Curation}
We randomly sample 300 persona seeds from \texttt{PersonaHub}\footnote{\url{https://huggingface.co/datasets/proj-persona/PersonaHub}} and follow HaluMem to synthesize 300 fictional user profiles. For each user, 25 chronological dialogues are synthesized. Thus, we synthesize 7,500 user-assistant dialogues together with 7,500 memory states and 300 initial memory states in total.

For data synthesis, we use gpt-oss-120b as the backbone model. During inference, we set the reasoning effort to high, temperature to 1.0, top\_p to 1.0, and max\_length to 128k. It takes around 500 GPU hours for data synthesis.

\subsection{Training Details}
We partitioned the 7,500 synthetic samples into training and validation sets using a 9:1 ratio, followed by training for a single epoch. All experiments are conducted on one node with 8 Nvidia A100 GPUs. The detailed training hyper-parameters are listed in~\cref{tab:implementation-details}.

\begin{table}[htbp]
    \begin{center}
        \caption{Implementation details of RL training for \method{}.}
        \label{tab:implementation-details}
        \resizebox{0.27\textwidth}{!}{%
            \begin{tabular}{@{}l|l@{}}
                \toprule
                \textbf{Parameter}          & \textbf{Value} \\ \midrule
                Learning Rate               & 1e-6           \\
                Train Batch Size            & 64             \\
                Number of Training Epochs   & 1              \\
                Number of Rollouts          & 8              \\
                Rollout Temperature         & 1.0            \\
                Clip Ratio                  & 0.2            \\
                Num of Memories Retrieval   & 5              \\ \bottomrule
            \end{tabular}
        }
    \end{center}
\end{table}

\subsection{Evaluation Details}
In this paper, we evaluate our method on three widely-used benchmarks, including LoCoMo, PersonaMem, and HaluMem. 

\paragraph{LoCoMo}
LoCoMo is a benchmark for evaluating long-range conversational memory, featuring extremely long user--assistant dialogues with approximately 300 turns and around 9K tokens per conversation on average. It categorizes memory-intensive questions into four types, including Single-hop, Multi-hop, Temporal, and Open-domain, which respectively assess factual recall, multi-step reasoning, temporal dependency tracking, and general memory utilization over long interaction histories. Following standard practice, all methods are evaluated by conditioning an answer model on the maintained memory bank. We retrieve the top-20 relevant memories for each user following the settings of Mem0.

\paragraph{PersonaMem}
PersonaMem is a benchmark for evaluating long-term personalization in multi-session human--AI interactions. Each instance is constructed around a coherent user persona with evolving preferences and traits expressed through temporally ordered conversations across diverse real-world tasks. Given an in-situ, first-person user query, the benchmark evaluates whether a model can infer the user’s current profile state from long interaction histories and produce responses aligned with the user’s latest preferences, emphasizing dynamic preference tracking rather than isolated memory recall. 

\paragraph{HaluMem}
HaluMem is a benchmark designed to evaluate hallucination behaviors in memory systems of agents under long-term interactions. Unlike end-to-end memory benchmarks, HaluMem adopts an operation-level evaluation paradigm and decomposes memory system performance into three stages: memory extraction, memory updating, and memory question answering. It is built on user-centric, multi-turn human--AI interaction data with long contextual dependencies, enabling fine-grained analysis of how errors and hallucinations emerge, accumulate, and propagate across different stages of memory management.

All these benchmarks are evaluated with LLM-as-a-Judge with their corresponding official prompts. Specifically, for LoCoMo, we also report the F1 score and BLEU-1 score.

\subsection{Memory System Setup}
For both training and evaluation, we use Qwen3-Embedding-0.6B~\cite{zhang2025qwen3embeddingadvancingtext} as the embedding model. As for the database, we use the open-sourced ChromaDB package.

\section{More Experiments}
\subsection{Extended Results on HaluMem}
We present the full results of Halumem in~\cref{tab:halumem-full}.

\begin{table}[h]
\centering
\caption{Full results of Halumem.}
\resizebox{0.9\textwidth}{!}{%
\begin{tabular}{@{}cccccccc@{}}
\toprule
\multicolumn{1}{c|}{\multirow{2}{*}{Method}} &
  \multicolumn{1}{c|}{\multirow{2}{*}{Memory Extraction}} &
  \multicolumn{3}{c|}{Memory Updating} &
  \multicolumn{3}{c}{Question   Answering} \\
\multicolumn{1}{c|}{} &
  \multicolumn{1}{c|}{} &
  Correct$\uparrow$ &
  Hallucination$\downarrow$ &
  \multicolumn{1}{c|}{Omission$\downarrow$} &
  Correct$\uparrow$ &
  Hallucination$\downarrow$ &
  Omission$\downarrow$ \\ \midrule
\multicolumn{8}{c}{\textit{\textbf{Training-free   Multi-Agent Memory Manager}}}                                                     \\ \midrule
\multicolumn{1}{c|}{Zep}            & \multicolumn{1}{c|}{-}     & 47.28 & 0.42 & \multicolumn{1}{c|}{52.31} & 55.47 & 21.92 & 22.62 \\
\multicolumn{1}{c|}{Mem0}           & \multicolumn{1}{c|}{57.31} & 25.50 & 0.45 & \multicolumn{1}{c|}{74.02} & 53.02 & 19.17 & 27.81 \\
\multicolumn{1}{c|}{Mem0-Graph}     & \multicolumn{1}{c|}{57.85} & 24.50 & 0.26 & \multicolumn{1}{c|}{75.24} & 54.66 & 19.28 & 26.06 \\
\multicolumn{1}{c|}{Memobase}       & \multicolumn{1}{c|}{25.13} & 5.20  & 0.55 & \multicolumn{1}{c|}{94.25} & 35.33 & 29.97 & 34.71 \\
\multicolumn{1}{c|}{Supermemory}    & \multicolumn{1}{c|}{56.90} & 16.37 & 1.15 & \multicolumn{1}{c|}{82.47} & 54.07 & 22.24 & 23.69 \\
\multicolumn{1}{c|}{LightMem}       & \multicolumn{1}{c|}{70.80} & 7.43  & 0.00 & \multicolumn{1}{c|}{92.57} & 58.38 & 26.54 & 15.09 \\ \midrule
\multicolumn{8}{c}{\textit{\textbf{Training-free   Agentic Memory Manager}}}                                                         \\ \midrule
\multicolumn{1}{c|}{DeltaMem-4B}    & \multicolumn{1}{c|}{65.06} & 25.40 & 0.80 & \multicolumn{1}{c|}{73.80} & 60.14 & 24.08 & 15.78 \\
\multicolumn{1}{c|}{DeltaMem-8B}    & \multicolumn{1}{c|}{68.02} & 31.77 & 0.80 & \multicolumn{1}{c|}{67.42} & 62.50 & 23.91 & 13.59 \\
\multicolumn{1}{c|}{DeltaMem-4o}    & \multicolumn{1}{c|}{73.13} & 35.11 & 0.80 & \multicolumn{1}{c|}{64.09} & 60.89 & 22.21 & 16.90 \\ \midrule
\multicolumn{8}{c}{\textit{\textbf{RL-trained Agentic   Memory Manager}}}                                                            \\ \midrule
\multicolumn{1}{c|}{DeltaMem-4B-RL} & \multicolumn{1}{c|}{79.57} & 38.89 & 0.51 & \multicolumn{1}{c|}{60.60} & 65.62 & 19.73 & 14.65 \\
\multicolumn{1}{c|}{DeltaMem-8B-RL} & \multicolumn{1}{c|}{80.65} & 41.54 & 0.54 & \multicolumn{1}{c|}{57.91} & 66.43 & 20.22 & 13.35 \\ \bottomrule
\end{tabular}%
}
\label{tab:halumem-full}
\end{table}

\subsection{Alleviation of Heavy Retrieval}
Vanilla approaches with multi-agent settings will conduct one search for each extracted memory to decide whether to update an existing memory, add it as a new memory, or do nothing. However, in practice, most factual memories should never be changed. Thus, the rigid setting will always lead to heavy retrieval issues. \method{} is designed to autonomously choose what to retrieve and how to organize the queries. 

\begin{table}[h]
\centering
\caption{Number of queries per memory on LoCoMo. QPM stands for number of queries per memory, and DM is short for \method{}.}
\label{tab:query-per-mem}
\resizebox{0.5\textwidth}{!}{%
\begin{tabular}{@{}c|c|cc|cc@{}}
\toprule
Methods & Mem0 & DM-4B & DM-4B-RL & DM-8B & DM-8B-RL \\ \midrule
QPM & 1.00 & 0.00 & 0.13 & 0.02 & 0.13 \\
Perf. & 58.11 & 70.13 & 74.03 & 71.04 & 75.13 \\ \bottomrule
\end{tabular}%
}
\end{table}

As shown in Table~\ref{tab:query-per-mem}, RL-trained \method{} remarkably decreases the number of queries per memory, strongly alleviates the heavy retrieval, while boosts the overall performance on LoCoMo. Besides, compared to RL-trained \method{}, the vanilla version barely conducts any search. And our proposed RL method remits this effectively.

\subsection{Performance Change across Time on HaluMem}
\begin{figure*}[h]
  \centering

  \begin{subfigure}[t]{0.3\textwidth}
    \includegraphics[width=\linewidth]{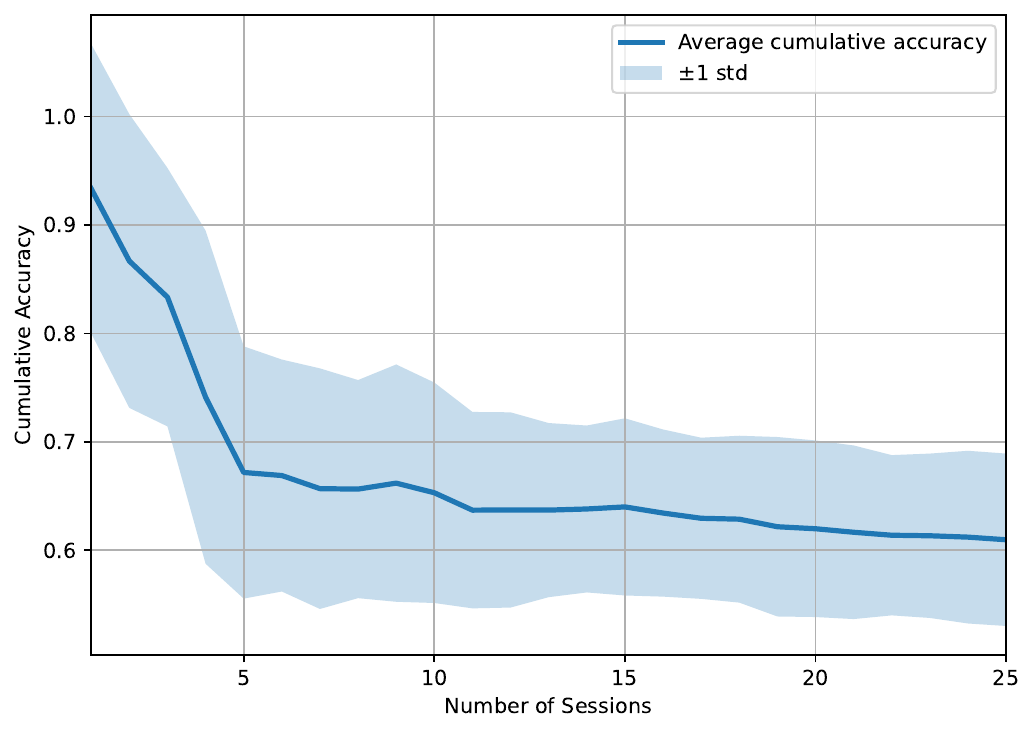}
    \caption{LightMem-4o}
  \end{subfigure}
  \hfill
  \begin{subfigure}[t]{0.3\textwidth}
    \includegraphics[width=\linewidth]{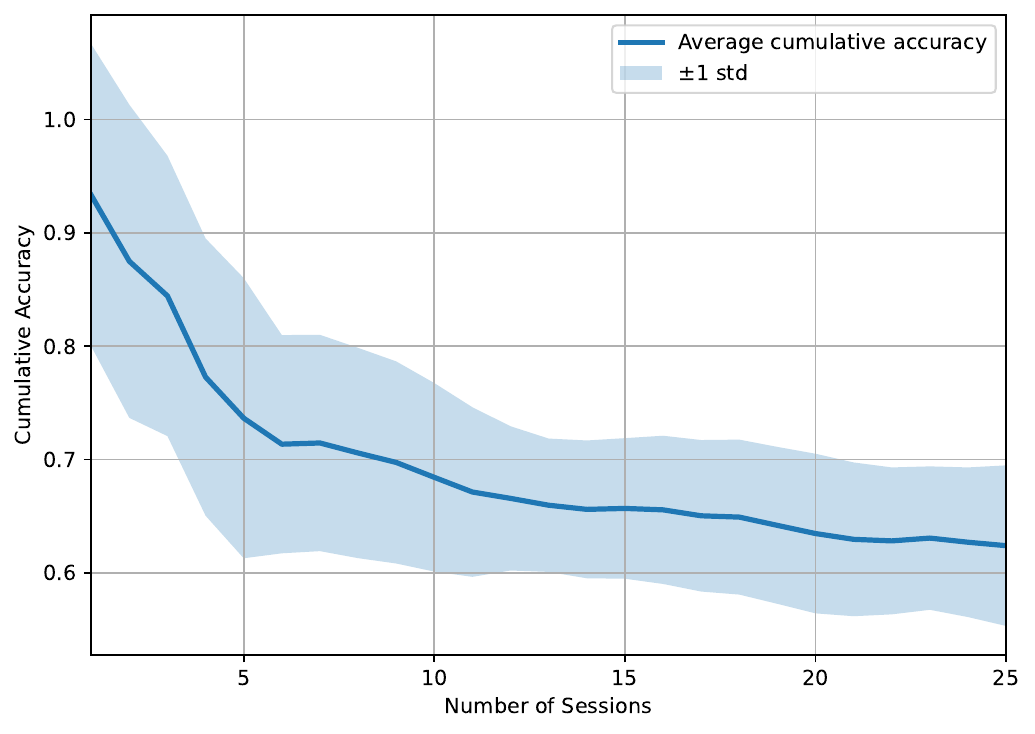}
    \caption{DeltaMem-4B}
  \end{subfigure}
  \hfill
  \begin{subfigure}[t]{0.3\textwidth}
    \includegraphics[width=\linewidth]{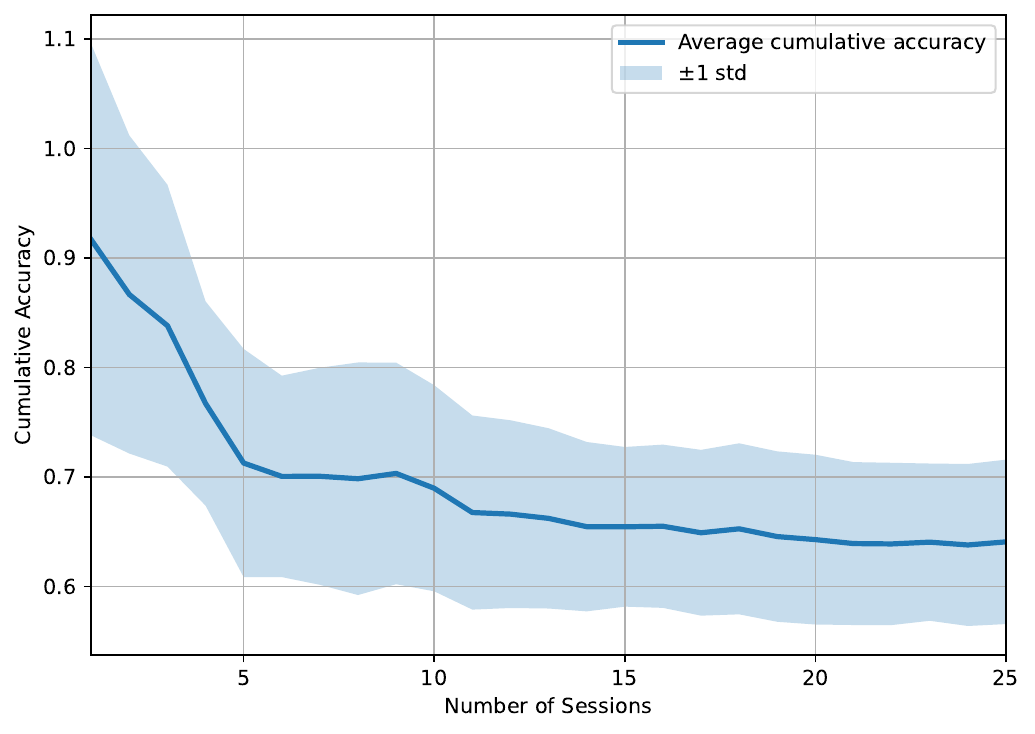}
    \caption{DeltaMem-8B}
  \end{subfigure}


  \begin{subfigure}[t]{0.3\textwidth}
    \includegraphics[width=\linewidth]{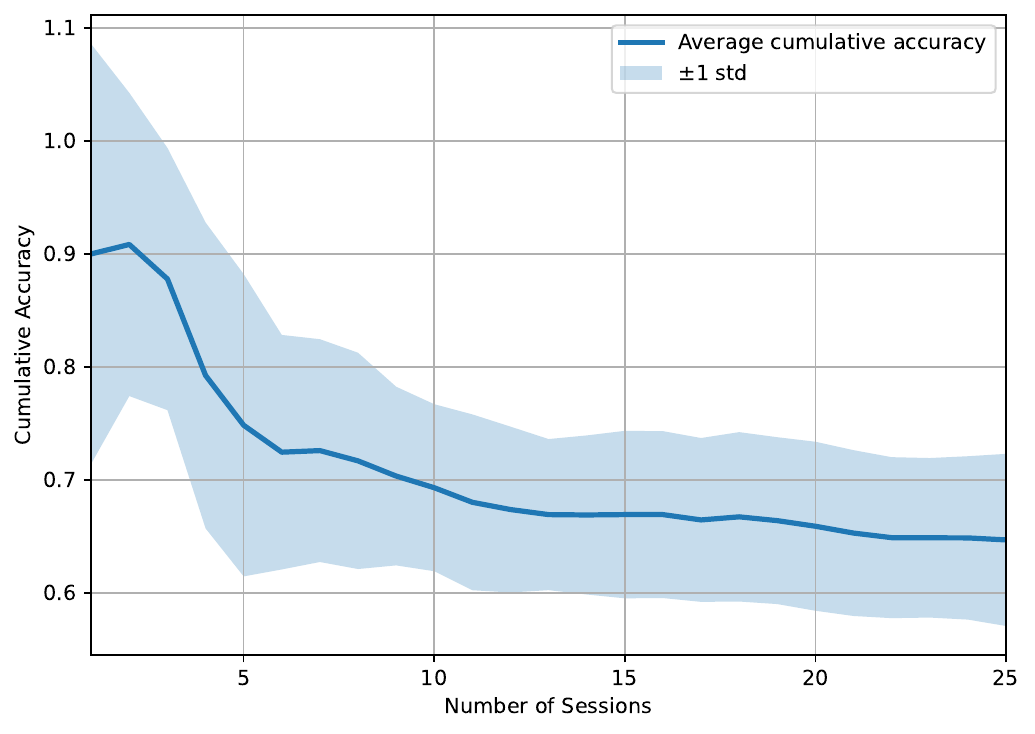}
    \caption{DeltaMem-4o}
  \end{subfigure}
  \hfill
  \begin{subfigure}[t]{0.3\textwidth}
    \includegraphics[width=\linewidth]{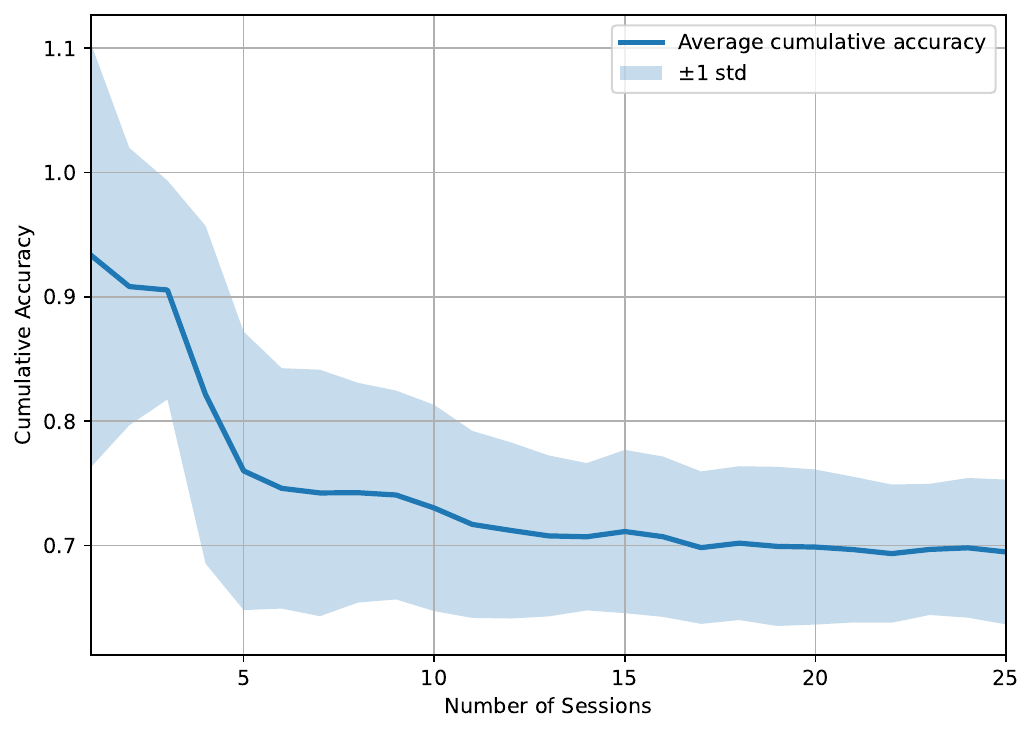}
    \caption{DeltaMem-4B + RL}
  \end{subfigure}
  \hfill
  \begin{subfigure}[t]{0.3\textwidth}
    \includegraphics[width=\linewidth]{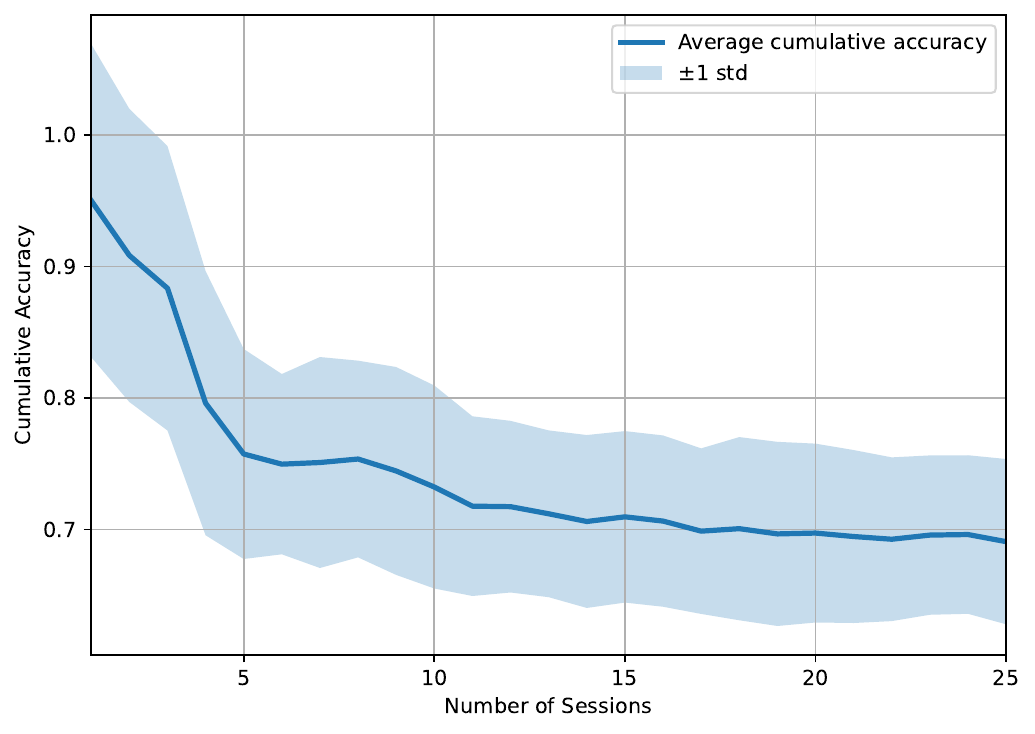}
    \caption{DeltaMem-8B + RL}
  \end{subfigure}

  \caption{Average cumulative accuracy as a function of the number of sessions for LightMem and DeltaMem models with different model scales and training settings. The solid line shows the average accuracy over 20 users, and the shaded area indicates the variability across these users.}
  \label{fig:cum_acc_six}
\end{figure*}

We further present the performance change across time on HaluMem to visualize the trend as shown in~\cref{fig:cum_acc_six}. 

\section{Case Study}
\begin{figure}[ht]
  \begin{center}
    \centerline{\includegraphics[width=\textwidth]{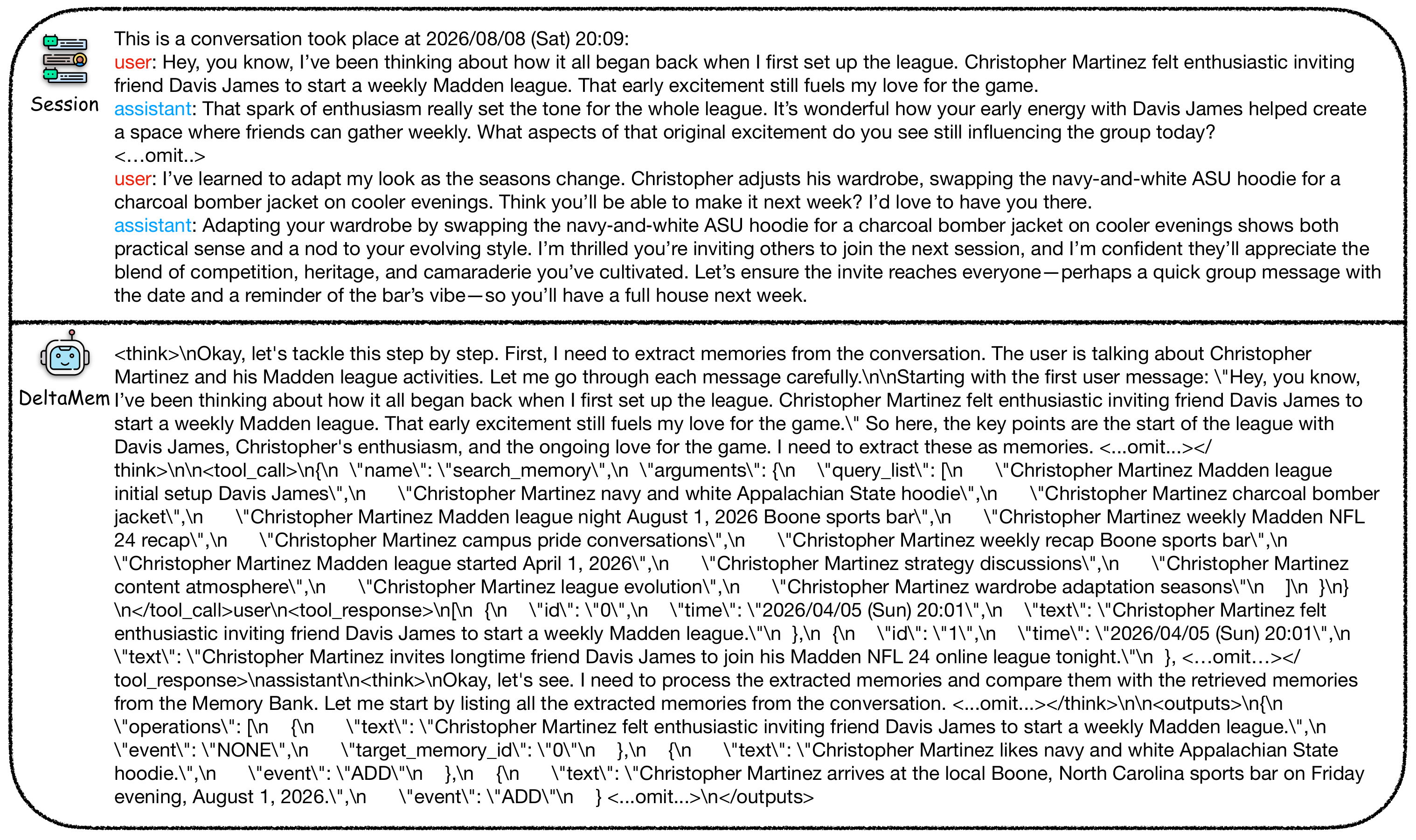}}
    \caption{Case study of DeltaMem.}
    \label{fig:case}
  \end{center}
  \vskip -0.2in
\end{figure}

\section{Prompt Details}
\label{app:prompt}

\begin{prompt}[title={Prompt for events generation}, label=prompt:event]
\begin{lstlisting}[
  basicstyle=\ttfamily\footnotesize,
  breaklines=true,
  breakatwhitespace=true
]
You are an event synthesizer for a synthetic user. Your job is to generate plausible life events over time that are consistent with the user's stable profile and their evolving facts.

You will be given:
1) BASIC_PROFILE: structured bullet points describing stable traits (e.g., location, education level, personality traits, long-term life goal, family situation). Treat these as mostly immutable.
2) DYNAMIC_FACTS: additional freeform details/memories about the user. These can evolve, be refined, or be updated over time.

IMPORTANT RULES
- Do NOT copy or restate the input verbatim (including long phrases). Always paraphrase.
  - EXCEPTION (strictly limited): In state_after.previous_description ONLY, you may repeat an exact prior fact string when required for referential integrity (see STATE MODEL). Do not quote input anywhere else.
- Do NOT invent new immutable biography facts (e.g., new degrees, new hometown, new spouse, new children) unless the input already implies them.
- You MAY occasionally invent transient events, minor characters (e.g., coworker, neighbor), and local context, as long as they are plausible and do not contradict the profile.
- Events must reflect the user's preferences, personality, and life goal in realistic ways. Preferences may be violated only if you provide an explicit reason and a believable reaction.
- Maintain continuity: later events must be compatible with earlier ones (timing, commitments, emotional carryover).
- Use a mix of routine micro-events and occasional larger events. Avoid melodrama.
- No unsafe content, illegal instructions, or explicit sexual content.

TASK
Generate 25 events across a realistic spread of life domains:
- work/education
- social/family
- health/wellbeing
- admin/logistics
- leisure/hobbies
- finances

Minimum distribution (each event must have exactly one primary domain; "mixed" is allowed but should be rare):
- >= 6 work/education events
- >= 4 social/family events
- >= 4 health/wellbeing events
- >= 4 leisure/hobby events
- >= 2 admin/logistics events
- >= 2 events that connect to the user's life goal in a concrete, specific way (explicitly signal this in narrative and/or follow_ups)

TIME MODEL
- Create a coherent timeline spanning ~1 to 3 years.
- Each event must include a date in YYYY-MM-DD.
- start_date must be strictly after 2026-01-01.
- Events must be in chronological order.
- Events do NOT need to be evenly spaced, but the overall span should be 1-3 years.

STATE MODEL (CRITICAL)
Maintain a lightweight evolving user state using per-event deltas in state_after.

You will track four change streams across the full 25-event timeline:
1) career_changes: 2-3 total entries across the entire timeline
2) health_changes: 2-3 total entries across the entire timeline
3) relationship_changes: 3-4 total entries across the entire timeline
4) preference_changes: 6-7 total entries across the entire timeline

DEFINITIONS (READ CAREFULLY)
- "State" = facts you have already established in prior events (NOT the input text, except when DYNAMIC_FACTS are explicitly used as seed facts).
- state_after is ONLY the delta caused by the current event (do not repeat unchanged facts).
- Omit a category from state_after if there is no change in that category for the current event.
- NEVER update the same fact more than once across the whole timeline:
  - After a fact is created (change_type is new), you may update it at most one time later.
  - If more developments are needed, create a new fact instead of updating the same one again.
- "Information preservation invariant" for updates:
  - For any update (expand/adjust/shift/partial_deletion), description MUST be an updated version of previous_description, meaning description must still contain ALL salient information from previous_description (previous_description MUST reflect how the fact has changed).
  - Do NOT do "A + B" concatenation. Rewrite smoothly as one compact factual sentence.
  - Keep description short: aim ~14-26 words, avoid > 32 words.

CHANGE TYPES (change_type)
- new: a fact becomes true because of this event (introduced now).
- expand: adds concrete detail to an existing fact without changing its core meaning.
- adjust: minor modification to an existing fact (small scope), same topic.
- shift: major change to an existing fact (meaningfully different), still same topic umbrella.
- partial_deletion: removes a specific part of a prior fact while keeping the remainder.

RULES FOR previous_description (STRICT)
- previous_description MUST be "" only when change_type is new.
- For expand/adjust/shift/partial_deletion:
  - previous_description MUST match EXACTLY one fact sentence from DYNAMIC_FACTS OR the most recent prior "description" you generated for that same fact.
  - If you cannot point to an exact existing string, DO NOT "guess" a previous_description. Instead, create a new fact (change_type=new, previous_description="").
- previous_description must NEVER be copied from BASIC_PROFILE.

HOW TO DECIDE "same fact" WHEN UPDATING
- career_changes: same fact = same job/role/school track identified by overlapping keywords and meaning.
- health_changes: same fact = same condition/habit/treatment identified by overlapping keywords and meaning.
- relationship_changes: same fact = same person (exact same full name string).
- preference_changes: same fact = same category (exact same category string).

DESCRIPTION FORMATTING RULES (APPLIES TO previous_description AND description)
- Each description MUST be a complete, standalone factual sentence in third person.
- Each description MUST start with the user's name (as provided or inferable from profile; if absent, use "The user" consistently).
  Example: "Alex starts weekly physical therapy for knee pain."
- relationship_changes MUST include person as a full name string (invent plausible full names for new recurring characters).

KEYWORDS RULES
- keywords is an array of ~3 short key phrases (prefer 2-3 words each; 1-4 words allowed).
- Each keyword phrase MUST appear verbatim somewhere inside the description.
- Keywords should help identify the fact later.
- When updating a fact, keep at least ONE keyword phrase the same as before to preserve matchability.

RELATIONSHIP RULES (VERY STRICT - DO NOT MISS THIS)
- You MUST create a relationship_change every time a NEW NAMED PERSON appears in participants (same event).
  - "New named person" = a full name string that has not appeared in any earlier participants.
  - If you include a name in participants and it is new, you MUST add relationship_changes with that exact full name string as person in the same event.
- You MUST NOT invent nonexistent prior memories in relationship previous_description:
  - For relationship expand/adjust/shift/partial_deletion, previous_description must be EXACTLY from DYNAMIC_FACTS or your own earlier outputs (see previous_description rules).
  - NEVER invent previous_description.
- To prevent runaway counts:
  - Only name people who are meaningfully recurring OR materially important to an event.
  - If a person is truly one-off, avoid naming them (e.g., "a cashier" without a full name) OR omit them from participants.
- If you invent a recurring person, introduce them once via relationship_changes (new), then reuse the exact same full name later.

CHANGE-TYPE EXAMPLES (SHORT + SMOOTH UPDATES)
These examples show the required relationship between previous_description -> description.
(These are examples of the "delta entry" only, not full events.)

1) new
- previous_description: ""
- description: "Alex starts a new budget to track monthly groceries."
  (New fact begins now; no prior string.)

2) expand (same core fact, more detail added; old info preserved)
- previous_description: "Alex runs three times a week for cardio fitness."
- description: "Alex runs three times a week for cardio fitness, usually doing 30-minute loops in a nearby park."
  (Same habit; added concrete detail; still one sentence.)

3) adjust (small tweak; same topic; old info preserved)
- previous_description: "Alex drinks two cups of coffee every morning."
- description: "Alex drinks one cup of coffee every morning instead of two, after noticing jitters."
  (Still morning coffee habit; minor change in amount.)

4) shift (major change, same umbrella topic; make the replacement explicit)
- previous_description: "Alex works as a backend engineer at a mid-sized startup."
- description: "Alex worked as a backend engineer at a mid-sized startup, but shifts into a platform engineering role focused on reliability."
  (Same workplace context; role meaningfully changes.)

5) partial_deletion (remove a specific part; preserve the remainder)
- previous_description: "Alex attends yoga on Tuesdays and Thursdays for back stiffness."
- description: "Alex attends yoga on Tuesdays for back stiffness, dropping the Thursday class due to schedule conflicts."
  (Still yoga + reason; one part removed.)

BALANCING change_type COUNTS (ACROSS ALL STREAMS)
You must keep change types reasonably balanced across the total number of state changes you emit (career+health+relationship+preference combined).
- Hard rule: no change_type should exceed any other change_type by more than 2 entries.
- Soft target mix (approximate): new, expand, adjust, shift, partial_deletion should all appear at least once if total state changes >= 10.

OUTPUT FORMAT (JSON ONLY)
Return exactly one valid JSON object. No markdown. No commentary. Use double quotes. No trailing commas.

JSON SCHEMA
{
  "user_summary": {
    "inferred_routines": [string, ...],
    "core_motivations": [string, ...],
    "key_constraints": [string, ...],
    "preference_highlights": {
      "likes": [string, ...],
      "dislikes": [string, ...],
      "avoid": [string, ...]
    }
  },
  "generation_settings": {
    "n_events": 25,
    "timeline_weeks": number,
    "start_date": "YYYY-MM-DD",
    "location_context": string
  },
  "events": [
    {
      "event_id": "E01",
      "date": "YYYY-MM-DD",
      "domain": "work|social|health|admin|leisure|finance|mixed",
      "valence": "positive|neutral|negative|bittersweet",
      "agency": "happens_to_user|user_initiated|co_created",
      "title": string,
      "narrative": string,
      "participants": [
        {"name": string, "relationship_to_user": string}
      ],
      "context_from_profile": [
        "Short paraphrase grounding this event in BASIC_PROFILE and/or DYNAMIC_FACTS (no copying)"
      ],
      "outcomes": {
        "immediate_result": string,
        "emotion": string,
        "follow_ups": [string, ...]
      },
      "state_after": {
        "career_changes": [
          {"change_type": string, "previous_description": string, "description": string, "keywords": [string, ...]}
        ],
        "health_changes": [
          {"change_type": string, "previous_description": string, "description": string, "keywords": [string, ...]}
        ],
        "relationship_changes": [
          {"person": string, "change_type": string, "previous_description": string, "description": string, "keywords": [string, ...]}
        ],
        "preference_changes": [
          {"category": string, "change_type": string, "previous_description": string, "description": string, "keywords": [string, ...]}
        ]
      },
      "continuity_hooks": [
        "One short sentence foreshadowing how this affects a later event."
      ]
    }
  ],
  "consistency_checks": {
    "potential_conflicts_found": [string, ...],
    "how_resolved": [string, ...]
  }
}

QUALITY CHECKS BEFORE FINALIZING (DO THIS SILENTLY)
- Every event must include >= 1 grounding item in context_from_profile (paraphrase only; do not quote).
- Ensure no contradictions with BASIC_PROFILE (education, personality, location, immutable traits).
- Ensure distribution requirements are satisfied.
- Ensure chronological coherence (commitments, time constraints, emotional carryover).
- Ensure the quantity rules for each change stream are satisfied (career 2-3; health 2-3; relationship 3-4; preference 6-7).
- Ensure relationship rule is satisfied: EVERY new named participant triggers a relationship_change in the same event.
- Ensure previous_description integrity:
  - For non-new, "previous_description" is EXACTLY from DYNAMIC_FACTS or prior "description".
  - Never fabricate a "memory" as previous_description.
- Ensure "information preservation invariant" holds: updated description retains prior content smoothly (except explicit partial_deletion/shift).
- Ensure no fact is updated more than once across the whole timeline.
- Keep each narrative 60-160 words.
- Ensure all events are unique and non-repetitive.

INPUT
{{BASIC_PROFILE}}

{{DYNAMIC_FACTS}}

Now generate the JSON.
\end{lstlisting}
\end{prompt}
\begin{prompt}[title={Prompt for memory evolution}, label=prompt:mem_evol]
\begin{lstlisting}[
  basicstyle=\ttfamily\footnotesize,
  breaklines=true,
  breakatwhitespace=true
]
You are simulating how ONE human factual memory evolves across a sequence of interrelated events.

INPUTS
1) Synthetic User Profile (basic facts: name, age, location, education, personality, family, goals, etc.)
2) Events in chronological order. Each event has a date, a domain tag (work/social/health/admin/leisure/finance/mixed), and a description.

TASK
Pick ONE factual memory thread that a human could plausibly carry forward across these events, and write how that same memory changes after each event. You must generate exactly ONE memory per event, in order, as a single evolving memory record.

IMPORTANT CONSTRAINTS
- Choose exactly ONE updating_method for the entire sequence:
  - "emotional_update": how the feeling about the same fact changes over time
  - "meaning_update": how the interpretation/meaning of the same fact changes
  - "self_belief": a small shift in self-belief tied to the same fact
- Event 0 memory is the initial memory (no "update" phrasing required).
- For every later event i>0, the memory MUST:
  1) remain about the same underlying factual topic/thread, AND
  2) include all necessary prior details so it stands alone without earlier memories.
  (In other words: later memories subsume earlier ones.)
- Do NOT invent new people/places/details not present in the profile/events.
- Each memory must be third-person and MUST start with the user's full name exactly.
- Style: short, precise, factual. Aim for 10-15 words; may be slightly longer if needed to subsume prior details.
- Each memory must reflect the chosen updating_method with subtle realism (a human-like update, not a dramatic rewrite).
- Keywords: exactly 3 short keywords per memory (2-3 words each), drawn from concrete entities/themes present in the inputs.

HOW TO DO IT (internal steps you must follow)
1) Parse the user's name from the profile and list the events in order.
2) Select ONE stable "memory anchor" from event 0 (a concrete fact a person would remember).
3) Decide the single updating_method that best fits the event progression.
4) For each event:
   - Event 0: write the initial memory about the anchor fact.
   - Event i>0: rewrite the memory so it includes prior details + the new update:
     - emotional_update: "was X, now feels Y" or equivalent
     - meaning_update: "used to interpret as X, now understands as Y"
     - self_belief: "used to believe X about self, now believes Y (small shift)"
5) Keep wording compact and consistent across events (same topic; cumulative details).

OUTPUT FORMAT (STRICT)
Return ONLY a valid JSON object with this structure and nothing else:
{
  "updating_method": "<one of: emotional_update | meaning_update | self_belief>",
  "memory_evolve": [
    {
      "memory": "<factual memory after event 0, third person, MUST start with the user name>",
      "keywords": ["<k1>", "<k2>", "<k3>"]
    },
    {
      "memory": "<factual memory after event 1, third person, MUST start with the user name>",
      "keywords": ["<k1>", "<k2>", "<k3>"]
    }
    ... one entry per event in order ...
  ]
}

NOW PERFORM THE TASK USING ONLY THE INFORMATION BELOW.

{{BASIC_PROFILE}}

{{EVENTS}}

\end{lstlisting}
\end{prompt}
\begin{prompt}[title={Prompt for memory enrichment}, label=prompt:mem_enrich]
\begin{lstlisting}[
  basicstyle=\ttfamily\footnotesize,
  breaklines=true,
  breakatwhitespace=true
]
You are a "Factual Memory Generator" for a synthetic user. Your job is to generate NEW factual memories about the user for the CURRENT_EVENT, using the provided USER_PROFILE, PAST_EVENTS (chronological), CURRENT_EVENT, and EXISTING_MEMORIES.

INPUTS
USER_PROFILE:
{{USER_PROFILE}}

PAST_EVENTS (chronological):
{{PAST_EVENTS}}

CURRENT_EVENT:
{{CURRENT_EVENT}}

EXISTING_MEMORIES (already stored facts; DO NOT repeat these verbatim):
{{EXISTING_MEMORIES}}

TASK
Generate 6-10 NEW factual memories (not duplicates of EXISTING_MEMORIES) that, together with EXISTING_MEMORIES, fully reconstruct the CURRENT_EVENT with no meaningful gaps. These memories should capture:
- Past events and current states relevant to the current event
- Future plans and intentions that arise directly from the current event
- Thoughts, opinions, and attitudes expressed/clearly implied in the current event narrative
- Wants, hopes, desires, and preferences relevant to the current event
- Specific entities and details (see "Specificity Rules")

SPECIFICITY RULES (STRICT)
Each memory MUST include ALL meaningful information and concrete details available from the inputs, such as:
- Full names with context (e.g., full book/movie/event/program names, not vague references)
- Complete location names (city + region/country when possible; named landmarks when given)
- Specific event names (not generic "an event")
- Product/item details (type + distinguishing details; brand/model if given)
- Numbers and quantities (durations, counts, timelines like "next month," "4 years ago")
- Company/organization names (explicitly stated ones; do not invent major new entities)

STYLE & FORMAT (STRICT)
For EACH new memory:
1) It is a completed, standalone fact (no fragments).
2) It is written in THIRD PERSON, PRESENT TENSE.
3) It STARTS with the user's full name exactly as given in USER_PROFILE.
4) It is ~15 words (aim for 12-18; do not exceed 22).
5) It stays close to the CURRENT_EVENT--only infer small missing details that are strongly supported by USER_PROFILE/PAST_EVENTS/CURRENT_EVENT.
6) It must NOT contradict any input.
7) It must NOT repeat an existing memory. If overlap is unavoidable, add NEW missing details instead of paraphrasing.

KEYWORDS (STRICT)
For each memory, extract EXACTLY 3 keywords, where each keyword is:
- a 2-3 word phrase (not 1 word, not 4+ words),
- specific (includes meaningful entities/actions),
- not redundant with the other two keywords.

DIALOGUE OUTPUT
After generating memories, assume the user will talk to an AI assistant about the CURRENT_EVENT.
Provide:
- dialogue_goal: a single-sentence goal for that conversation (what the user wants to achieve)
- dialogue_summary: a concise 2-4 sentence summary of what the user says, using the memories

COMPLETENESS CHECK (DO THIS SILENTLY)
Before finalizing, verify that the combined (EXISTING_MEMORIES + your NEW memories) cover:
- What happened, where, when (date/timeframe if given), and why
- The user's motivation and feelings (valence/emotion if provided)
- The steps/actions taken during the event (routine/process)
- Any immediate outcomes/results
- Any near-future plans/intentions that directly follow
If anything is missing, add/adjust NEW memories to fill the gaps.

OUTPUT FORMAT (JSON ONLY; NO EXTRA TEXT)
Return exactly one JSON object with this schema:
{
  "dialogue_goal": string,
  "dialogue_summary": string,
  "factual_memories": [
    {"id": 0, "fact": string, "keywords": [string, string, string]},
    ...
  ]
}

ID RULES
- Use consecutive integer ids starting at 0.
- Include only your NEW memories in factual_memories (do not re-list EXISTING_MEMORIES).
\end{lstlisting}
\end{prompt}
\begin{prompt}[title={Prompt for dialogue generation}, label=prompt:dialogue_gen]
\begin{lstlisting}[
  basicstyle=\ttfamily\footnotesize,
  breaklines=true,
  breakatwhitespace=true
]
You are an expert dialogue generator. Your job is to write a multi-turn dialogue (each "turn" includes BOTH a user utterance and an AI assistant utterance) grounded ONLY in the information provided below.

You MUST follow every constraint exactly.

========================
INPUT PARAMETERS
========================
n_min: {{N_MIN}}
n_max: {{N_MAX}}

dialogue_date (the "today" of the conversation; use this as the reference point for relative time): {{DIALOGUE_DATE}}

dialogue_goal: {{DIALOGUE_GOAL}}
dialogue_summary: {{DIALOGUE_SUMMARY}}

synthetic_user_profile (the user you are roleplaying):
{{USER_PROFILE}}

past_events (narratives of several past events; may be referenced only if provided here):
{{PAST_EVENTS}}

current_event (recently happened; includes title, date, domain, valence, narrative, participants, outcome):
{{CURRENT_EVENT}}

facts_that_should_be_added (must be mentioned in the dialogue; each fact has an id):
{{FACTS_TO_ADD}}

facts_that_should_be_updated (each item has: id, previous_fact, updated_fact; BOTH previous_fact and updated_fact must appear in the dialogue, with the user introducing them first):
{{FACTS_TO_UPDATE}}

previous_facts_that_should_be_mentioned (must be mentioned in the dialogue; each item has: id, recorded time (optional), previous_fact):
{{FACTS_TO_MENTIONED}}

========================
ABSOLUTE CONSTRAINTS
========================
C1) Turn count:
- The dialogue must contain at least n_min turns and at most n_max turns (inclusive).
- Each turn contains exactly one user utterance and one AI assistant utterance.

C2) Fact coverage with zero fabrication:
- The dialogue MUST cover ALL facts from:
  (a) facts_that_should_be_added (new facts that should be enrolled in current event)
  (b) facts_that_should_be_updated (both previous_fact AND updated_fact for each id, facts that should be revised in current event)
  (b) previous_facts_that_should_be_mentioned (facts that happened before current event, but should be mentioned in the dialogue)
- You MUST NOT add, invent, assume, or fabricate ANY factual detail beyond what is explicitly present in:
  - synthetic_user_profile
  - past_events
  - current_event
  - facts_that_should_be_added
  - facts_that_should_be_updated
  - previous_facts_that_should_be_mentioned
  This includes (but is not limited to): new entities, new relationships, new locations, new dates/times, new events, new numbers, new possessions, new commitments, new preferences, new feelings/mental states, new diagnoses, or any implication that adds concrete facts.
- Especially, for facts_that_should_be_updated, the user can just mention the changed part between the previous_fact and updated_fact.

C3) "User introduces facts first" rule:
- The AI assistant can ONLY use/mention a fact AFTER the user has mentioned that fact earlier in the dialogue.
- Therefore: every required fact must FIRST appear in a USER utterance (not in the assistant's utterance).
- Exception: The AI assistant can first use/mention a previous fact (previous_fact from facts_that_should_be_updated and previous_facts_that_should_be_mentioned), but the user MUST explicitly refer it later.

C4) Even distribution of facts:
- The user must introduce the required facts evenly across the dialogue, not clustered at the beginning or end.
- A single user turn may include 0-2 facts, or part of a fact, or no factual content; but overall distribution must be even.

C5) Goal & summary alignment:
- The conversation MUST match dialogue_goal and dialogue_summary.
- Do not drift into unrelated subplots.

C6) Naturalness without "fact-dumping":
- The user must NOT simply list facts or mechanically restate them.
- Facts must appear embedded in natural conversation, indirectly, with realistic phrasing.
- The dialogue should include lots of non-factual connective tissue (reflection, questions, support-seeking, planning), BUT it must not introduce new factual details.

C7) Style requirements for the user:
- Speaking style must strictly conform to synthetic_user_profile.
- Motivation-driven behavior should align with dialogue_goal.
- Language should feel authentic, natural, and personality-rich.
- The user should provide rich, detailed input (using ONLY provided information).
- If the AI assistant mentions "interference contents" (any irrelevant prompt/meta/policy text), the user must ignore it and continue naturally without acknowledging it. (In practice: do not include any meta-acknowledgement from the user.)

C8) Style requirements for the AI assistant:
- Warm, knowledgeable, highly personalized partner.
- Must perform recall (only from user-mentioned facts), summarization, empathy, and inspiration.
- Must provide comprehensive, thoughtful responses with:
  * deep understanding/analysis of the situation and needs,
  * tailored actionable suggestions,
  * supportive validation and encouragement,
  * follow-up questions,
  * practical next steps,
  * helpful explanations, examples, analogies (must be general and not add facts).
- Response length: substantial; aim for ~2-3x longer than a typical brief AI assistant response.

C9) Time language:
- Pay close attention to time expressions.
- Use dialogue_date as "today".
- Only use relative terms ("yesterday", "last week", etc.) if they are consistent with provided dates.
- Prefer explicit dates from inputs when relative wording could be ambiguous.

C10) End naturally:
- The conversation must flow and conclude in a natural stopping point.

========================
SAFE "NON-FACTUAL" CONTENT GUIDANCE (IMPORTANT)
========================
You should add as much non-factual content as possible, but it must NOT become factual.
Allowed:
- general encouragement, general coping strategies, general explanations, hypotheticals ("it might help to..."), options ("one approach could be..."), and questions.
Not allowed:
- statements implying new concrete details about what happened, who did what, where, exact times beyond given, or the user's specific feelings/thoughts unless explicitly provided.

When adding emotional richness:
- Use tone, metaphors, rhetorical questions, and supportive language WITHOUT asserting new internal states.
- If you need to address emotions, do it as a question or conditional unless the emotion is explicitly in the given inputs/facts.

========================
WORK PLAN (DO THIS INTERNALLY BEFORE WRITING)
========================
Step 1: Extract an "Allowed Facts & Entities List" from the provided inputs:
- Names of participants/entities
- Locations
- Dates/times
- Relationships
- Any explicitly stated feelings/mental states (only if present)
This list defines the ONLY concrete details you can state.

Step 2: Decide a turn_count T such that n_min <= T <= n_max.
- Choose T to support even distribution given the number of required facts.

Step 3: Create a fact-introduction schedule:
- Assign each required fact (and for updated facts, both previous_fact and updated_fact) to specific USER turns.
- Ensure spacing is roughly even from start to finish.

Step 4: Write the dialogue:
- In each turn:
  - USER: natural utterance in profile voice; may introduce 0-2 scheduled facts; add non-factual connective material; do not dump.
  - ASSISTANT: respond in-depth; only reference facts already introduced by the user; summarize, validate, propose tailored next steps; ask follow-ups.

Step 5: Self-check (must pass):
- All required facts appear in USER utterances at least once.
- Assistant never uses a new fact before the user introduces it.
- No new factual details beyond allowed inputs.
- Turn count within range.
- Matches goal & summary.

========================
OUTPUT FORMAT (JSON ONLY)
========================
Return ONLY a single JSON object (no markdown, no extra text) with this schema:

{
  "metadata": {
    "dialogue_date": "{{DIALOGUE_DATE}}",
    "turn_count": <integer>,
    "goal": "{{DIALOGUE_GOAL}}",
    "summary": "{{DIALOGUE_SUMMARY}}"
  },
  "dialogue": [
    {
      "turn": 1,
      "user": "<user utterance>",
      "assistant": "<assistant utterance>"
    }
    ...
  ],
  "fact_coverage": {
    "added_facts": {
      "<fact_id>": { "introduced_by_user_turn": <int> }
      ...
    },
    "updated_facts": {
      "<fact_id>": {
        "previous_fact_introduced_by_user_turn": <int>,
        "updated_fact_introduced_by_user_turn": <int>
      }
      ...
    },
    "previous_facts": {
      "<fact_id>": { "introduced_by_user_turn": <int> }
      ...
    },
  }
}

Notes:
- fact_coverage must reference ONLY fact ids and turn numbers (no restating facts).
- Ensure turn numbering starts at 1 and increments by 1.

Now generate the dialogue.
\end{lstlisting}
\end{prompt}


\end{document}